\documentclass[a4paper,fleqn,authoryear]{cas-sc}

\usepackage[authoryear,round]{natbib}

\usepackage{lineno}
\usepackage{algorithm}
\usepackage{algpseudocode}
\usepackage{amsmath}
\usepackage{float}
\usepackage{placeins}
\usepackage{graphicx}
\usepackage{subcaption}
\usepackage{multirow}
\usepackage{tablefootnote}
\def\tsc#1{\csdef{#1}{\textsc{\lowercase{#1}}\xspace}}
\tsc{WGM}
\tsc{QE}
\begin{document}
%\linenumbers % Enable line numbering

\ExplSyntaxOn
\cs_set:Npn \__first_footerline:
{
  \group_begin:
  \small
  \sffamily
  \ifnum\theblind>0\relax
  \else
    \__short_authors: :~
  \fi
  { \rmfamily \itshape Preprint }
  \group_end:
}
\ExplSyntaxOff

\let\WriteBookmarks\relax

% Short title
\shorttitle{FedDRAW}
% Short authors
\shortauthors{M. Moradpour et al.}

% Main title of the paper
\title [mode = title]{FedDRAW: Federated Dual Reputation Annealing Weighting for Heterogeneous Multi-Institutional Chest Radiograph Classification}

\author[label1]{Maryam Moradpour}[orcid=0009-0001-0906-5842] %% Author name

\author[label1,label2]{Anne-Christin Hauschild}[orcid=0000-0002-7499-4373]
\affiliation[label1]{organization={Institute for Predictive Deep Learning in Medicine and Healthcare, Justus-Liebig University, Gießen, Germany}}
\affiliation[label2]{organization={Institute for Medical Informatics, University Medical Center Göttingen, Germany}}

\begin{abstract}
Artificial intelligence models are promising for medical diagnosis, but 
they require large numbers of unbiased data, which in medicine are
distributed across hospitals and cannot be centralized to protect patient privacy. Federated Learning (FL) addresses this, since hospitals train one shared diagnostic model while patient data remain local. Training proceeds in communication rounds, in which each hospital trains the shared model locally and returns it to the server for merging by weighted average. This aggregation weight determines whose institutional knowledge shapes the result. Federated averaging (FedAvg) sets it in proportion to local sample count, so a small but informative hospital is permanently assigned a small influence, and largest clients could dominate the global model even when they are less informative. We propose Federated Dual Reputation Annealing Weighting (FedDRAW), a server-side aggregation method that combines a data-size prior with the cosine similarity between client and global parameters under two coupled annealing schedules. An inner schedule shifts client reputation from the size prior towards similarity. An outer, deferred annealing schedule on the softmax inverse temperature keeps the weighting selective in the early and middle rounds and relaxes it to uniformity at convergence. We evaluate FedDRAW on 12 simulated client-partition scenarios of two chest radiograph datasets (CheXpert and ChestMNIST), against seven federated baselines under identical local training settings. FedDRAW achieved the highest average rank among all eight methods under both AUC and the geometric mean (GM) of sensitivity and specificity, which a Friedman test with Nemenyi post-hoc analysis confirmed to be a statistically significant difference between the methods. Scheduling two signals, rather than fixing the weights by sample count alone, could enable less biased diagnostic models.
\end{abstract}

\begin{highlights}
    \item FedDRAW: federated learning that schedules aggregation influence across rounds instead of fixing it by hospital size
    \item Client weights shift from data size to model similarity during training
    \item A deferred schedule returns all client weights to equal at convergence
    \item Small hospitals are not permanently outweighed by larger ones
    \item Best average rank compared to eight state-of-the-art methods on twelve chest X-ray client partitions
\end{highlights}


\maketitle

\section{Introduction}
\label{sec:intro}

Machine learning, and deep learning in particular, has shown promising results for
clinical tasks that include diagnosis, risk prediction, and treatment
planning~\citep{rajpurkar2022ai}, and medical image analysis is one of the areas in which
deep neural networks have been applied most extensively~\citep{litjens2017survey}. Such models generally require
large numbers of annotated cases, and the training data should cover the variation that occurs across
the patient populations on which the model is later used. A single institution may not satisfy either requirement, because its cohort could reflect one catchment population, one set of scanners and acquisition protocols, and one
annotation practice; hence, a model trained at one location might carry the bias of that location and perform worse elsewhere ~\citep{arasteh2023enhancing,li2021fedbn}. Conventional centralized training assumes that the data required to build such models can be assembled in one location. In healthcare, this assumption rarely holds. Clinical data
remain distributed across hospitals and medical centers and cannot readily be pooled
because of patient privacy, legal and regulatory constraints, institutional
data-governance policies, data ownership, and security considerations~\citep{kairouz2021advances,guan2024medical}. Privacy, however, is not the only obstacle. Clinical data are naturally distributed among institutions; 
an individual site often holds a dataset that is small, imbalanced, or unrepresentative of the wider patient population, and a locally trained model therefore generalizes poorly~\citep{guan2024medical}. 
Transferring large clinical archives to a single center may also be impractical. In multi-institutional
chest-radiograph analysis, for example, collaborative training has been shown to improve
diagnostic performance on an unseen institution relative to single-site training, with the
largest on-domain benefit observed for institutions holding smaller local
datasets~\citep{arasteh2023enhancing}.

Federated learning (FL)~\citep{mcmahan2017communication} was introduced as a collaborative
learning paradigm in which participating institutions train a shared model while their raw
data remain local~\citep{hauschild2022federated,park2024federated,moradpour2026federatedrsf}. 
FL has accordingly been taken up across multi-institutional healthcare applications, with
medical imaging among the most active~\citep{guan2024medical}. Therefore, in this study we use 
public medical imaging datasets as examples. The data held by participating institutions, however, are not
statistically identical. Heterogeneity in FL is not a single phenomenon but comprises
feature-distribution skew, label-distribution skew, concept drift, concept shift, and
quantity skew~\citep{kairouz2021advances}. In clinical imaging, these forms occur together,
because institutions differ in the number of patients they serve, in disease prevalence
and class proportions, in scanners, acquisition protocols and contrast settings, in
patient populations, and in annotation protocols and the interpretation of ambiguous
findings~\citep{li2021fedbn,liu2026adaptive,arasteh2023enhancing}. FL settings are commonly
distinguished by how these data are partitioned across participants, and three categories
are conventionally identified, namely horizontal FL, vertical FL, and federated transfer
learning~\citep{yang2019federated,kairouz2021advances}. In horizontal FL, the participants
hold different samples that are described by the same features and address the same
learning task. In vertical FL, they hold complementary feature subsets for a largely
overlapping set of samples, and each participant therefore optimizes a different part of
the model. In federated transfer learning, the participants differ in both their samples
and their feature spaces, and knowledge is transferred between the resulting heterogeneous
domains rather than merged into one shared model~\citep{yang2019federated}.

This distinction determines what the server is able to aggregate, and horizontal FL is the
relevant category for the present work. Because horizontally partitioned clients share a
feature space and a prediction task, they can train an identical network architecture
whose parameters are dimensionally and semantically comparable, and local models and their
differences can therefore be compared and merged directly at the
server~\citep{shi2025fedawa,rehman2023ldawa}. This study adopts a horizontal cross-silo
setting, in which healthcare institutions hold different patient cohorts and
collaboratively train a compatible global model. A communication round then proceeds in
four steps. The server distributes the current global model to the participating clients,
each client trains that model on its local data, the clients return their locally updated
parameters, and the server aggregates the received models to form the next global
model~\citep{mcmahan2017communication}. Because heterogeneous clients produce different
local models, the server must decide how strongly each client should affect the result. We
refer to this quantity as the aggregation weight, that is, the relative influence of a
client's local update on the next global model, and it is the component of the federated
pipeline that this work modifies.

Federated Averaging (FedAvg)~\citep{mcmahan2017communication} is the standard reference for
this step. The server averages the returned local models with weights proportional to the
number of local training samples; hence the aggregation criterion is the client sample
count and nothing else. The rule is inexpensive, requires no information beyond quantities
the server already knows, and reflects the reasonable expectation that larger local
datasets yield statistically more stable local estimates. Its criterion, however, remains
a function of sample quantity throughout training, and several studies have argued that
this quantity is an incomplete indicator of a client's value. Dataset size does not reveal
how representative a client's class distribution is of the intended global
objective~\citep{ye2023feddisco}; larger local datasets do not necessarily produce more
reliable updates, and size-based weighting can amplify poorly trained or distributionally
biased client models~\citep{kundroo2026fedlbw}. Because weight is tied to volume, methods of
this type judge a client by the amount of data it holds rather than by how informative that
data is, so a small institution may retain persistently limited influence even when its
updates are valuable~\citep{zeng2021fedcav}.
This concern is not only theoretical, as increasing
the sample counts of clients whose local models perform poorly has been shown to
destabilize size-weighted averaging while leaving distance-based weighting
stable~\citep{yeganeh2020ida}.

Subsequent FedAvg-based methods addressed related but distinct difficulties.
FedProx~\citep{li2020fedprox} constrains excessive local deviation through a proximal
penalty on the local objective and admits partial work from slower devices, targeting
statistical and systems heterogeneity together, whereas
SCAFFOLD~\citep{karimireddy2020scaffold} corrects client drift by adjusting each local
gradient with server and client control variates, at the cost of persistent per-client
state. Meanwhile, FedNova~\citep{wang2020fednova} addresses the objective inconsistency
caused by unequal local computation, normalizing each client's accumulated update by its
local progress, hence clients performing more local steps do not acquire disproportionate
influence. These methods improve local optimization stability, drift correction, and
update normalization, yet none of them reconsiders the criterion by which client
importance is assigned. FedNova is instructive in this respect, because it restores the
intended sample-size objective rather than asking whether that objective distributes
influence appropriately. Conventional FedAvg-based approaches therefore improve federated
optimization under heterogeneous conditions while leaving unresolved the persistent
dominance of data-size-based influence and the limited contribution assigned to small but
potentially informative clients.

A natural response is to replace or supplement the sample count with a signal derived from
the client models themselves, and model similarity has become the most widely used such
signal because it relates local client knowledge to the evolving global model. An early
and influential step in this direction is FedAdp~\citep{wu2021fedadp}, which measures the
angle between each inferred local gradient and a sample-size-weighted estimate of the
global gradient and converts a smoothed version of this angle into contribution scores
through a decreasing nonlinear mapping and a softmax. The same work, however, also exposed
a weakness of the signal itself. The angle is highly random at the beginning of federated
training, and FedAdp restrains this instability by averaging the angle over its history, a
remedy that still lets similarity act on aggregation from the very first round, when the
smoothed value coincides with the instantaneous one. Later methods refined where and how
similarity is measured. L-DAWA~\citep{rehman2023ldawa}, developed for federated
self-supervised representation learning, moves the comparison to the level of individual
layers and scales each layer's contribution by its angular alignment with the previous
global model, while SimProx~\citep{elniss2024simprox} combines cosine and Gaussian
similarity after finding a single metric incomplete, at the price of a pairwise matrix
whose cost grows quadratically with the number of clients.
FedSim~\citep{palihawadana2022fedsim} instead turns similarity into a clustering criterion,
grouping clients by their gradients in every round partly to prevent a large group of
similar clients from dominating the global update, although the benefit of this clustering
depends on the heterogeneity regime. A further question concerns what similarity actually
rewards. FedSaC~\citep{yan2024fedsac} shows that collaboration among the most similar
clients may add no information, whereas moderate differences between client feature
distributions can supply complementary knowledge, and personalized federated methods have
been designed explicitly to avoid the negative transfer that collaboration with unrelated
clients can induce~\citep{xu2023fedpac}.
A client that diverges from
the current global model may hold rare presentations, underrepresented populations, or unusual
acquisition conditions; hence, low similarity should not by itself reduce its contribution
to a negligible level. Taken together, these studies establish similarity as an informative signal, yet none of them adjusts the trust placed in that signal according to the stage of training, even though the signal is measured against a global model that is itself still forming in the early rounds.

Similarity is not the only information that has been incorporated into the weights.
FedDisco~\citep{ye2023feddisco} augments the size prior with a measure of how far each
client's label distribution lies from a target distribution, although this discrepancy is
computed once before training and the resulting weights are reused across all rounds.
FedLBW~\citep{kundroo2026fedlbw} goes further and discards the sample count altogether,
weighting clients by the inverse validation loss of their models on a class-balanced
dataset held at the server. However, the need for such a labeled evaluation set at the
server becomes a concern, because the weights then depend on how representative that set
is. FedLAW~\citep{li2023revisiting} shares this dependence, learning the client weights and
a global shrinking factor on a server-side proxy dataset. Other approaches let the weights
emerge from the models themselves. FedAWA~\citep{shi2025fedawa} optimizes the weights of
each round in favor of updates aligned with the merged global direction, and
FedLaw~\citep{lu2025fedlaw} scores each client by the agreement of its update with those of
the other clients, smoothed over the client's contribution history. Two observations
follow. First, although the numerical weights of these methods differ between rounds, they
differ because the client state has changed and not because the weighting rule itself is
scheduled. The scoring or optimization criterion is applied identically in every round,
and none of these formulations contains a coefficient that varies deliberately with
training progress, and the relative trust placed in each signal therefore remains fixed
even though the information those signals carry is not equally reliable early and
late~\citep{ye2023feddisco,shi2025fedawa,li2023revisiting,lu2025fedlaw}. Second, adaptation
does not by itself distribute influence. A criterion that rewards agreement with the
majority direction can repeatedly favor the same clients, because alignment-based
weighting reduces the influence of updates that differ from the merged
direction~\citep{shi2025fedawa}, and contribution defined through similarity to other
clients treats conformity with the collective update pattern as the measure of
value~\citep{lu2025fedlaw}.

Three limitations therefore remain. Data-size weighting is stable and immediately
available, yet it systematically restricts the contribution of small but informative
clients. Model similarity carries information that sample count cannot express, yet the one method that examines its early behavior reports that its alignment signal is full of randomness at the beginning of training and restrains this by averaging over history rather than by adapting the influence of the signal to its maturity~\citep{wu2021fedadp}. We further hypothesize that one contributing factor is the immaturity of the evolving global reference against which the signal is measured. Unconstrained dynamic weighting, finally, may repeatedly
favor large, highly similar, or majority-aligned clients, leaving rare or complementary
institutional knowledge with little opportunity to shape the global model. Existing
methods thus do not provide a stage-aware aggregation mechanism that exploits the
stability of data quantity early, progressively incorporates model similarity as the
global model matures, and at the same time controls how strongly aggregation influence may
concentrate. This motivates the question addressed here, namely how federated aggregation
can coordinate data quantity and model similarity according to their changing reliability
while preserving a meaningful contribution opportunity for heterogeneous clients.

Annealing offers a principled way to introduce a component gradually rather than at full
strength from the first iteration. The idea originates in simulated annealing, where the
system is allowed to move to a high-risk state with high probability during an initial
high-temperature stage in order to escape local optima, and this exploration is
progressively reduced as the temperature
falls~\citep{kirkpatrick1983simulated,nguyen2021gradual}. In federated learning, the
mechanism has so far served either as a search procedure over training configurations such
as participant subsets, hyperparameters, and quantization
schemes~\citep{neto2022fedsa,li2024fedfq,peleg2025pause}, or as a client-side schedule that
gradually shifts each client's reliance from its local model toward the global model as
that model matures~\citep{nguyen2021gradual}. In all of these, the server-side aggregation
rule itself remains unchanged. Notably, the client-side transition rests on the
observation that the averaged global model is not necessarily better than local models in
the early stage of training~\citep{nguyen2021gradual}, an observation that concerns the very model against which client-to-global similarity is measured, and which, together with the stability of the data-size prior from the first round, suggests that the relative importance of the two criteria should follow the training stage rather than remain fixed.

We accordingly propose Federated Dual Reputation Annealing Weighting (FedDRAW), a
server-side aggregation method that combines client data size with client-to-global model
similarity and anneals their roles across communication rounds without modifying
client-side training. The term dual denotes two coupled schedules. An inner schedule
governs the mixture of the two criteria, anchoring the early rounds in the data-size prior
and smoothly increasing the contribution of similarity as the global reference matures. An
outer schedule governs the temperature at which the resulting client scores are converted
into aggregation weights, and it is annealed in a deferred manner, preserving a
discriminative weight distribution in the early and middle rounds and relaxing it toward a
uniform distribution in the final rounds. The two schedules address the identified
limitations jointly. The inner schedule matches each criterion to the stage of training in
which it is reliable, while the outer schedule controls concentration, because a mechanism
that only increased the influence of similarity would risk transferring dominance from the
largest clients to the most globally aligned ones.

Fairness carries different meanings across federated learning studies, ranging from a more
uniform distribution of predictive performance across clients to parity between
demographic groups and to rewards proportional to measured contribution, and we review
these notions in Section~\ref{subsec:rw-fairness}. In this work, fairness refers to
aggregation-level equal treatment, meaning that every client participating in a
communication round receives the same server-side aggregation coefficient in the later
rounds, irrespective of its local dataset size or the alignment of its update. This is a
statement about server-assigned coefficients alone, not a claim of equal predictive
performance across clients, of demographic parity, of contribution-proportional reward, or
of equal selection frequency, and because local updates may differ in magnitude, equality
of coefficients is not equality of numerical influence. Relaxing the weights toward
uniformity also raises the question of whether a poorly performing client could then harm
the global model. In our setting, every client begins each round from the shared global
model and clients are assumed to be honest rather than adversarial; hence, their updates
are refinements of the current global model rather than arbitrary or corrupted
contributions, and equal late-round coefficients do not admit harmful updates by
construction; robustness to adversarial participants is outside the scope of this work.

We evaluate FedDRAW on multi-label chest-radiograph classification using CheXpert and on
ChestMNIST, under heterogeneous client partitions, and compare it with representative
conventional, adaptive, and similarity-based aggregation methods. The contributions of
this work are as follows.

\begin{itemize}
  \item \textbf{Dual-criterion server-side aggregation.} We propose FedDRAW, a
  server-side aggregation method that jointly considers client data quantity and
  client-to-global model similarity when determining client influence, requiring no change
  to client-side training and no proxy dataset at the server.

  \item \textbf{Stage-aware inner annealing.} We introduce a schedule that progressively
  reduces reliance on the data-size prior and increases the contribution of model similarity as federated training advances, introducing the signal gradually rather than at full strength from the first round, in contrast to a fixed combination of the two criteria.

  \item \textbf{Outer annealing toward uniform weighting.} We introduce a second, outer
  schedule on the aggregation temperature that relaxes the weight distribution received
  from the inner annealing toward uniformity in the later rounds, and we define the
  resulting property as aggregation-level equal-treatment fairness, distinguishing it from
  performance, group, and contribution fairness.

  \item \textbf{Empirical validation on medical imaging.} We evaluate FedDRAW on twelve
  heterogeneous federated partitions of CheXpert and ChestMNIST against seven
  conventional, adaptive, and similarity-based aggregation baselines under identical local
  hyperparameters.

  \item \textbf{Statistical evaluation.} We compare the eight methods across all partition
  scenarios with per-metric Friedman tests followed by post-hoc analysis, and we
  additionally examine the behavior of the weighting mechanism itself, including the
  measured heterogeneity of the client partitions and the distribution of aggregation
  weights across clients and rounds.
\end{itemize}

The remainder of this paper is organized as follows.
Section~\ref{sec:related} reviews related work on federated optimization, aggregation
weighting, fairness, and federated learning in medical imaging.
Section~\ref{sec:method} presents the datasets and the FedDRAW method, including the two
annealing schedules and their configuration. Section~\ref{sec:results} describes the
experimental setup, the client partitions, the baselines, and the evaluation protocol, and
reports the experimental results together with the statistical analysis.
Section~\ref{sec:conclusion} concludes the paper and outlines directions for future work.

% ============================================================
% FedDRAW — Related Work
% ============================================================

\section{Related Work}
\label{sec:related}

This section reviews prior work according to the component of the federated pipeline that
each method modifies, since that is the axis along which the present contribution is
positioned. Section~\ref{subsec:rw-optimization} covers methods that act on local
optimization, which improve how client updates are produced while leaving the rule that
combines them intact. Section~\ref{subsec:rw-weighting} turns to methods that modify the
aggregation rule itself by deriving client influence from information other than the local
sample count. Section~\ref{subsec:rw-fairness} surveys the incompatible senses in which
fairness has been used in federated learning and states which of them the property claimed
for FedDRAW does and does not correspond to. Section~\ref{subsec:rw-medical} situates these
developments in multi-institutional medical imaging, the setting in which FedDRAW is
evaluated, and examines the evaluation metrics adopted in that literature. Throughout, each
subsection closes by identifying what the reviewed methods leave unresolved with respect to
the assignment of client influence.

\subsection{Federated Optimization under Heterogeneous Client Data}
\label{subsec:rw-optimization}

Federated learning must account for statistical and systems heterogeneity, because clients
may differ in data distributions, dataset sizes, computational resources, local update
counts, and participation patterns~\citep{kairouz2021advances,li2020fedprox}. Within this
setting, FedAvg~\citep{mcmahan2017communication} established the standard iterative
formulation in which participating clients perform local optimization and the server
aggregates their models using sample-size-proportional weights. Under heterogeneous client
objectives, however, local optimization may produce client drift or objective
inconsistency, which can slow convergence and degrade global-model
performance~\citep{karimireddy2020scaffold,wang2020fednova}.

Several established methods address heterogeneity by modifying local optimization.
FedProx~\citep{li2020fedprox} introduces proximal regularization to constrain local models
toward the current global model and tolerates partial work from slower devices, whereas
SCAFFOLD~\citep{karimireddy2020scaffold} uses server and client control variates to correct
client-specific update directions at the cost of persistent per-client state. Similarly,
FedDyn~\citep{acar2021feddyn} dynamically regularizes the local objectives so that local
stationary points are better aligned with the stationary point of the global objective,
while MOON~\citep{li2021moon} applies model-level contrastive learning, using the similarity
between the representations of the current local model, the global model, and the previous
local model to correct local training under heterogeneous client data. The central
intervention of these methods is the generation of local updates rather than the
construction of a round-dependent rule for client influence. It is worth noting that MOON
also uses representation similarity, although as a local regularizer rather than as a
server-side weighting signal. FedDRAW leaves the local training objective unchanged and acts
only on the weights assigned to the resulting client models, so the two families of methods
are compatible rather than competing. By contrast, FedNova~\citep{wang2020fednova} occupies
an intermediate position, since it normalizes each client's accumulated update according to
its effective local progress, so that clients performing more local optimization steps do
not acquire disproportionate influence, thereby restoring the sample-size objective that
unequal local computation would otherwise distort. This correction protects the intended
weighting rather than reconsidering whether that rule assigns influence appropriately, which
is the question FedDRAW addresses.

\subsection{Aggregation Weighting beyond Sample Count}
\label{subsec:rw-weighting}

A second line of work modifies the aggregation rule itself by using client information
beyond the local sample count. For instance, FedDisco~\citep{ye2023feddisco} combines
relative dataset size with a scalar discrepancy between each client's category distribution
and a target distribution, so that larger datasets and lower discrepancies receive higher
weight. It is conceptually close to FedDRAW in that both challenge pure sample-size
weighting, although the discrepancy depends on the fixed local label distribution and is
therefore determined before training and reused across communication rounds. FedDRAW
instead uses a signal that is recomputed from the current client and global models and, more
importantly, changes the relative influence of the two signals as training progresses.

Some methods derive client influence from model geometry or from measured performance.
IDA~\citep{yeganeh2020ida} weights each client by the inverse distance between its
parameters and the mean model of the participating cohort, and reports that increasing the
sample counts of poorly performing clients destabilizes size-weighted averaging while
leaving distance-based weighting stable. Similarly, FedAdp~\citep{wu2021fedadp} measures the
angle between each inferred local gradient and a sample-size-weighted estimate of the global
gradient, maps a smoothed angle through a decreasing nonlinear function, and normalizes the resulting contribution scores with a softmax, and it is the most directly related adaptive-weighting baseline evaluated in this work. When the participating clients differ in dataset size, the FedAdp weight retains a multiplicative sample-size factor alongside the contribution score, so the method combines quantity with alignment rather than replacing the former by the latter~\citep{wu2021fedadp}. FedDRAW also combines these two signals, and differs in that the relative influence of each is scheduled across communication rounds rather than fixed by a single expression. It is also the only method reviewed here
that reports an early-training difficulty with its own signal, since the angle between the
local and estimated global gradient is highly random at the beginning of training and is
therefore replaced by a cumulative smoothed angle. Although this restrains the instability,
it does so through historical averaging rather than through a stage-dependent activation of
the signal, since the smoothed value coincides with the instantaneous angle at the first
round and the signal influences aggregation from the outset. On the other hand,
FedLAW~\citep{li2023revisiting} learns a global shrinking factor together with the relative
client weights by optimizing a loss on a server-side proxy dataset, and
FedLBW~\citep{kundroo2026fedlbw} replaces sample-count weighting entirely with normalized
inverse validation losses obtained by evaluating each uploaded model on a labeled
server-side proxy set. Both therefore make the representativeness of the proxy set a
determinant of the weights, whereas FedDRAW requires no server-side data. For instance,
FedAWA~\citep{shi2025fedawa} forms a client vector as the difference between each returned
local model and the incoming global model and optimizes the weights of the round so that
vectors aligned with the merged global direction receive greater influence, while
FedDRW~\citep{li2025feddrw} retains a sample-size component but modulates it with a softmax
over the similarity between each client's feature representation and its label-distribution
cluster centre, for long-tailed federated classification. Server-side methods may also
exploit client diversity through model integration rather than through scalar reweighting,
as in FedEL~\citep{wu2024fedel}, which first obtains a conventional federated model and
subsequently constructs a server-side ensemble from client-specific classifier modules. This
changes the structure of the deployed model, whereas FedDRAW retains a single shared
architecture and modifies only the aggregation coefficients.

Across the methods surveyed in this subsection, the numerical weights may differ between
rounds, although they differ because the client state has changed rather than because the
weighting rule itself is scheduled. The scoring or optimization criterion is applied
identically in every round, and none of these formulations contains a coefficient that
varies deliberately with training
progress~\citep{ye2023feddisco,shi2025fedawa,li2023revisiting,kundroo2026fedlbw}. A second
common characteristic is that the criteria these methods adopt tend, by construction, to
reduce the influence of clients whose updates diverge from the prevailing consensus, since
weight is granted for proximity to a cohort mean~\citep{yeganeh2020ida}, for alignment with
an estimated global direction~\citep{wu2021fedadp,shi2025fedawa}, or for agreement with a
target distribution~\citep{ye2023feddisco}. A small institution holding rare presentations or
an underrepresented population is therefore liable to receive persistently low influence
precisely because its data are distinctive, and the reviewed methods do not treat the
preservation of such contributions as an explicit design objective. Although these methods
adapt client influence according to the current state or measured behavior of the clients,
FedDRAW differs in two respects. It schedules the composition of the client score across
communication rounds rather than applying a stationary criterion, and it separately
schedules the concentration of the resulting weight distribution so that no client retains
disproportionate influence over the final model.

\subsection{Fairness Notions in Federated Learning}
\label{subsec:rw-fairness}

Fairness has been used in federated learning for several incompatible objectives, which
differ in which quantity is equalized and for whom. Client-level performance fairness
balances predictive performance across participating clients, whether by up weighting
clients with higher losses~\citep{li2020fair}, by measuring the dispersion of client
accuracy~\citep{li2021ditto}, by optimizing against the worst-case mixture of client
distributions~\citep{mohri2019agnostic}, by seeking an update direction that increases no
participant's loss~\citep{hu2022fedmgda}, or by penalizing the spread of losses across
clients~\citep{yue2023gifair}. Group fairness instead equalizes prediction-rate
disparities between demographic groups defined by a sensitive attribute, adjusting the
aggregation weights from client-reported fairness statistics~\citep{ezzeldin2023fairfed}.
Collaborative fairness allocates model benefit in proportion to measured contribution and
therefore regards identical treatment of unequal contributors as
unfair~\citep{lyu2020collaborative,lu2025fedlaw}, while a fourth line equalizes how often
clients are selected rather than how their updates are weighted once
selected~\citep{peleg2025pause}.

Equal aggregation weights are consequently not what most of this literature means by
fairness, since several of these methods make weights deliberately unequal in order to
equalize outcomes~\citep{li2020fair,wang2024fedeba}. The property claimed for FedDRAW is
distinct and procedural. We use aggregation-level equal treatment to denote the condition
in which every client participating in a communication round receives the same server-side
aggregation coefficient in the later rounds, irrespective of its dataset size or the
alignment of its update, which prevents any client from retaining disproportionate
influence over the final model. The closest precedent is
FedEBA$+$~\citep{wang2024fedeba}, which likewise limits the concentration of the weight
vector, although it ties the weight itself to client underperformance and applies its
entropy term as a fixed regularizer, whereas FedDRAW derives the weight from data size and
client-to-global similarity and schedules the permitted concentration across communication
rounds. The claim concerns server-assigned coefficients alone. It is not a claim of equal
predictive performance across clients, of demographic parity, of contribution-proportional
reward, or of equal selection frequency, and because local updates may differ in
magnitude, equality of coefficients is not equality of numerical influence.

\subsection{Federated Learning in Medical Imaging}
\label{subsec:rw-medical}

Medical imaging is an important application of federated learning, because clinically
relevant data are often distributed across institutions and difficult to pool for reasons of
privacy, governance, and data-sharing constraints~\citep{guan2024medical}. Institutional
datasets may differ in scanner hardware, acquisition protocols, preprocessing procedures,
patient populations, disease prevalence, and annotation practices, producing feature, label,
and domain shifts~\citep{guan2024medical,arasteh2023enhancing,li2021fedbn}. Moreover, in
some federations the annotated label sets themselves do not fully
overlap~\citep{kulkarni2024surgical}.

A multi-institutional chest-radiograph study reports that on-domain performance was
influenced primarily by the amount of training data, whereas off-domain performance
benefited more strongly from greater training-data diversity in collaborative
learning~\citep{arasteh2023enhancing}. This distinction between volume and diversity is
directly relevant to aggregation design, since a weighting rule tied to volume alone rewards
the former and not the latter. FedBN~\citep{li2021fedbn} addresses a related feature-shift
problem by retaining batch-normalization parameters locally while aggregating the remaining
model parameters. These methods adapt what is shared or how local training proceeds, rather
than redefining the relative aggregation coefficient assigned to each participating client,
which is the question FedDRAW addresses.

Work on chest radiography has converged on a small number of benchmarks. CheXpert provides
224{,}316 radiographs from 65{,}240 patients labeled for fourteen
observations~\citep{irvin2019chexpert}, whereas ChestMNIST provides a standardized
lower-resolution multi-label chest-radiograph task within the MedMNIST
collection~\citep{yang2023medmnist}. Both are framed as multi-label thoracic-pathology
classification, and both are substantially imbalanced at the label level, since several
findings are present in only a small fraction of studies. Evaluation on these benchmarks is
dominated by the area under the receiver operating characteristic curve. The MedMNIST
benchmark protocol reports the area under the curve together with
accuracy~\citep{yang2023medmnist}, cross-institutional federated evaluation on chest
radiographs is reported in terms of the area under the
curve~\citep{arasteh2023enhancing,kulkarni2024surgical}, and comparisons between supervised
and label-efficient methods on the CheXpert test set are likewise reported as a mean area
under the curve averaged over the competition pathologies~\citep{tiu2022expert}.

Both metrics respond weakly to the class imbalance that characterizes these datasets, and
this limitation has been examined directly in the medical-informatics literature. Optimizing
and reporting a global area under the curve implicitly assumes that all decision thresholds
are equally relevant, whereas clinicians operate within ranges of acceptable thresholds that
reflect the differing consequences of false positives and false
negatives~\citep{cabitza2026why}. Accuracy, for its part, can be artificially inflated under
class imbalance, and threshold-dependent metrics are frequently computed at an arbitrary
cut-off that rarely corresponds to an actual decision trade-off~\citep{cabitza2026why}. The
consequence is that a favorable value under threshold-free evaluation does not establish
that an operating point exists at which an infrequent finding is detected at a clinically
acceptable rate, so reported performance may be optimistic relative to what a deployed system
would achieve once a threshold is fixed~\citep{cabitza2026why}. The point is practical rather
than formal, since a system supporting a clinical decision necessarily operates at a chosen
threshold, and the findings whose detection motivates such a system are frequently the
infrequent ones.

For this reason we complement the macro-averaged area under the curve with the geometric mean
of sensitivity and specificity, computed at per-class thresholds selected on a validation
split and then held fixed for test evaluation. Selecting condition-specific thresholds on a
validation set that does not overlap the test set is established practice on
CheXpert~\citep{tiu2022expert}. The geometric mean is small whenever either sensitivity or
specificity is small, so it cannot be inflated by strong performance on the majority class
alone, and reporting it alongside the area under the curve characterizes performance both
across all thresholds and at the single operating point that a deployed system would use.

% ============================================================
% Method Section
% ============================================================
\section{Method}
\label{sec:method}

In this section, we present the proposed method in full detail. We first
introduce the two chest radiograph datasets on which our study is built,
since their characteristics motivate several design decisions of the
method. We then formalize the federated learning setting and the
aggregation problem, and introduce FedDRAW, a server-side aggregation
framework built on two coupled annealing mechanisms, namely an inner
annealing that governs the informational basis of client reputation,
and an outer annealing that governs how reputation is translated into
aggregation weights. Each component of the framework is described in a
dedicated subsection, followed by the complete algorithm.

\subsection{Datasets}
\label{sec:datasets}
 
We evaluate the proposed FedDRAW framework on two publicly available
multi-label chest radiograph datasets, CheXpert~\citep{irvin2019chexpert}
and ChestMNIST~\citep{yang2023medmnist}. The two datasets differ in
size, image resolution, patient population, and, importantly, in the
automated labeling pipelines from which their annotations were derived;
consistent behavior across both therefore indicates that the evaluated
aggregation strategies are not tied to the characteristics of a single
data source or label-extraction process. For consistency across
experiments, we consider the same five thoracic findings in both
datasets, namely Atelectasis, Edema, Pleural Effusion, Cardiomegaly,
and Consolidation. Figure~\ref{fig:dataset_examples} shows
representative test-set examples from both datasets together with their
binary labels for the five findings.

\vspace{5pt}

\begin{center}

\refstepcounter{table}
\label{tab:train_distributions}

\noindent\parbox{\textwidth}{%
\scriptsize
\textbf{Table \thetable.} Label distribution of the training data used in
this study for (a)~CheXpert and (b)~ChestMNIST. Each count gives the
number of images for which the corresponding finding is positive. Since
the task is multi-label, an image may be positive for several findings
simultaneously, so the counts are not mutually exclusive and do not sum
to the total number of images. The final row of each subtable gives the
number of images that are negative for all five findings.
}

\vspace{5pt}

\begin{minipage}[t]{0.48\textwidth}
\centering

{\scriptsize
\textbf{(a)} CheXpert (191,027 frontal radiographs used, after uncertainty
mapping).
\par}

\vspace{3pt}

\footnotesize
\begin{tabular}{lr}
\toprule
Class & Positive images \\
\midrule
Atelectasis          & 59,583 \\
Edema                & 61,493 \\
Pleural Effusion     & 76,899 \\
Cardiomegaly         & 23,385 \\
Consolidation        & 12,983 \\
All five labels $=0$ & 51,052 \\
\bottomrule
\end{tabular}

\end{minipage}
\hfill
\begin{minipage}[t]{0.48\textwidth}
\centering

{\scriptsize
\textbf{(b)} ChestMNIST (official training split of 78,468 images, used
in its entirety).
\par}

\vspace{3pt}

\footnotesize
\begin{tabular}{lr}
\toprule
Class & Positive images \\
\midrule
Atelectasis          & 7,996 \\
Edema                & 1,690 \\
Pleural Effusion     & 9,261 \\
Cardiomegaly         & 1,950 \\
Consolidation        & 3,263 \\
All five labels $=0$ & 59,438 \\
\bottomrule
\end{tabular}

\end{minipage}

\end{center}

\vspace{5pt}

\paragraph{CheXpert.}
CheXpert is a large-scale chest radiograph dataset containing 224,316
chest X-ray images from 65,240 patients~\citep{irvin2019chexpert}. Its
labels were extracted automatically from radiology reports by a
rule-based labeler and take one of three values per finding, namely
positive, negative, or uncertain, the last of which reflects ambiguous
report language. The dataset provides annotations for fourteen
observations, which its authors selected on the basis of their
prevalence in the reports and their clinical
relevance~\citep{irvin2019chexpert}. Among these fourteen, the CheXpert
authors designated five observations, namely Atelectasis, Cardiomegaly,
Consolidation, Edema, and Pleural Effusion, as the competition tasks of
the associated benchmark, selected according to their clinical
importance and prevalence~\citep{irvin2019chexpert}. We adopt these same
five findings as our prediction targets. For label preprocessing, the
five target observations were converted to binary labels. Following the
uncertainty-handling strategy of Pillai~\citep{pillai2022multi},
uncertain labels ($-1$) for Atelectasis and Edema were mapped to the
positive class (U-Ones), whereas uncertain labels for Cardiomegaly,
Consolidation, and Pleural Effusion were mapped to the negative class
(U-Zeros); missing label entries were likewise assigned to the negative
class, yielding binary labels for all five findings. In this study, we
work with the 191,027 frontal chest radiographs of the dataset. The
selected five pathologies exhibit substantial differences in
prevalence, providing a naturally imbalanced multi-label setting, since
Pleural Effusion and Edema are among the more prevalent findings,
whereas Consolidation is considerably less frequent. Images for which
none of the five selected findings is positive are retained as negative
samples for the considered label set.
Table~\ref{tab:train_distributions}(a) reports the resulting label
distribution of the CheXpert training data used in this work.
\vspace{4pt}

\begin{center}
\setlength{\tabcolsep}{3pt}
\renewcommand{\arraystretch}{1.0}

\begin{tabular}{ccccc}
% ---------- Row 1: CheXpert images ----------
\includegraphics[width=0.185\textwidth]{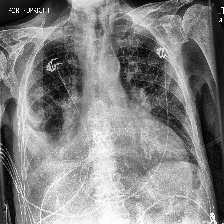} &
\includegraphics[width=0.185\textwidth]{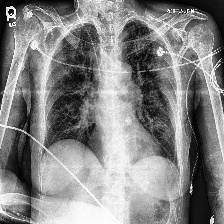} &
\includegraphics[width=0.185\textwidth]{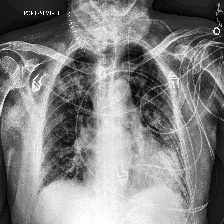} &
\includegraphics[width=0.185\textwidth]{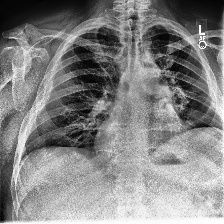} &
\includegraphics[width=0.185\textwidth]{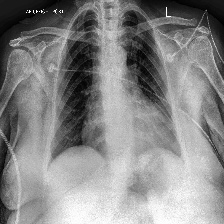} \\
% ---------- Row 2: CheXpert labels ----------
{\scriptsize\shortstack[l]{Atelectasis: 1\\Edema: 0\\Pleural Effusion: 1\\Cardiomegaly: 1\\Consolidation: 1}} &
{\scriptsize\shortstack[l]{Atelectasis: 0\\Edema: 1\\Pleural Effusion: 0\\Cardiomegaly: 0\\Consolidation: 0}} &
{\scriptsize\shortstack[l]{Atelectasis: 0\\Edema: 1\\Pleural Effusion: 0\\Cardiomegaly: 1\\Consolidation: 0}} &
{\scriptsize\shortstack[l]{Atelectasis: 0\\Edema: 0\\Pleural Effusion: 0\\Cardiomegaly: 0\\Consolidation: 0}} &
{\scriptsize\shortstack[l]{Atelectasis: 0\\Edema: 0\\Pleural Effusion: 0\\Cardiomegaly: 1\\Consolidation: 0}} \\[4pt]
\multicolumn{5}{c}{\footnotesize (a) CheXpert} \\[8pt]

% ---------- Row 3: ChestMNIST images ----------
\includegraphics[width=0.185\textwidth]{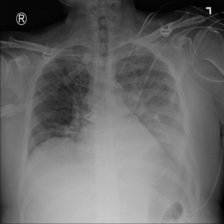} &
\includegraphics[width=0.185\textwidth]{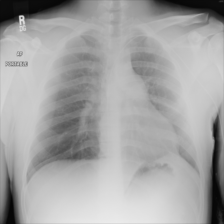} &
\includegraphics[width=0.185\textwidth]{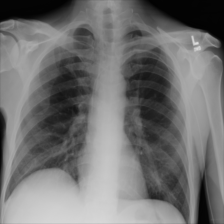} &
\includegraphics[width=0.185\textwidth]{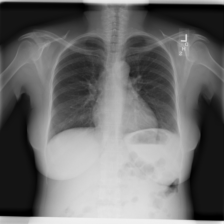} &
\includegraphics[width=0.185\textwidth]{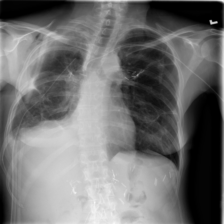} \\
% ---------- Row 4: ChestMNIST labels ----------
{\scriptsize\shortstack[l]{Atelectasis: 1\\Edema: 0\\Pleural Effusion: 0\\Cardiomegaly: 0\\Consolidation: 1}} &
{\scriptsize\shortstack[l]{Atelectasis: 1\\Edema: 0\\Pleural Effusion: 1\\Cardiomegaly: 0\\Consolidation: 1}} &
{\scriptsize\shortstack[l]{Atelectasis: 0\\Edema: 0\\Pleural Effusion: 1\\Cardiomegaly: 0\\Consolidation: 0}} &
{\scriptsize\shortstack[l]{Atelectasis: 0\\Edema: 0\\Pleural Effusion: 1\\Cardiomegaly: 1\\Consolidation: 0}} &
{\scriptsize\shortstack[l]{Atelectasis: 0\\Edema: 0\\Pleural Effusion: 0\\Cardiomegaly: 0\\Consolidation: 0}} \\[4pt]
\multicolumn{5}{c}{\footnotesize (b) ChestMNIST} \\
\end{tabular}

\vspace{4pt}

\refstepcounter{figure}
\label{fig:dataset_examples}

\noindent\parbox{\textwidth}{%
\scriptsize
\textbf{Figure \thefigure.} Representative test-set examples from
CheXpert~(a) and ChestMNIST~(b) with their binary labels for the five
considered findings. For the CheXpert test set, the reference labels
were obtained from radiologist annotations, with the majority vote of
five board-certified radiologists serving as the ground
truth~\citep{irvin2019chexpert}. The ChestMNIST labels originate from
the automated text mining of radiology reports in the source
ChestX-ray14 dataset~\citep{yang2023medmnist,wang2017chestxray}, and its
Effusion category is reported here under the harmonized name Pleural
Effusion for consistency with CheXpert.
}

\end{center}

\vspace{6pt}

\paragraph{ChestMNIST.}
ChestMNIST is the chest X-ray subset of the MedMNIST benchmark and is
derived from the NIH ChestX-ray14
dataset~\citep{yang2023medmnist,wang2017chestxray}, whose labels were
obtained through automated text mining of radiology reports. The source
dataset provides annotations for fourteen thoracic findings; from these,
we retain the same five pathologies used for CheXpert, providing a
consistent prediction task across the two datasets. The label
nomenclature of the two datasets differs slightly for one of the
selected findings, since the ChestX-ray14 label set, which ChestMNIST
inherits, denotes pleural fluid accumulation as Effusion, whereas
CheXpert uses the term Pleural Effusion. We treat these two categories
as corresponding to the same finding and refer to it throughout the
paper as Pleural Effusion.
 
In contrast to CheXpert, the labels distributed with ChestMNIST are
already binary, and each finding is annotated as either present ($1$) or
absent ($0$), with no uncertain category, since the task is formulated
by the benchmark providers as multi-label binary-class
classification~\citep{yang2023medmnist}. No additional uncertainty
handling is therefore required on our side, which also means that any
label noise present in ChestMNIST stems entirely from the automated
report-mining pipeline of its source dataset rather than from our
preprocessing choices. We use the official training split of the
benchmark in its entirety, without any further image selection. Like
CheXpert, ChestMNIST exhibits
substantial class imbalance, with positive findings representing only a
fraction of the available images for several pathologies; this
imbalance, combined with the smaller image resolution, makes ChestMNIST
a complementary and more challenging counterpart to CheXpert for
evaluating aggregation behavior. Table~\ref{tab:train_distributions}(b)
reports the label distribution of the ChestMNIST training data used in
this work. The construction of federated clients from these centralized
datasets is described in Section~\ref{sec:scenarios}.

\subsection{Preliminaries and Problem Formulation}
\label{sec:prelim}

Federated learning (FL) is a distributed learning paradigm in which
multiple clients collaboratively train a shared model under the
coordination of a central server, without ever exchanging their raw
data~\citep{mcmahan2017communication,kairouz2021advances}. Its central
goal is to obtain a global model that benefits from the collective data
of all participants while respecting data locality, a property of
particular importance in medical imaging, where patient data cannot
leave the acquiring institution. In each communication round, clients
train the current model on their private data and return only the
updated parameters; the server then combines these local models into a
new global model through weighted aggregation. The aggregation weights
determine how strongly each client influences the global model, and
this influence is decisive for both convergence and final performance,
and assigning weights that reflect each client's actual contribution
has been shown to accelerate convergence under heterogeneous
data~\citep{wu2021fedadp}. The de facto standard aggregation rule,
FedAvg~\citep{mcmahan2017communication}, weights each client by its
dataset size. As we argue below, this choice embeds two assumptions
that become problematic under realistic data heterogeneity. Addressing
these two limitations at the aggregation level is the starting point of
the proposed method.

We consider a federated learning system of $K$ clients indexed by
$i \in \{1,\dots,K\}$, each holding a private local dataset $D_i$ of
size $n_i = |D_i|$, with total data volume $N = \sum_{i=1}^{K} n_i$.
All clients share an identical neural network architecture; in this
work a DenseNet backbone~\citep{huang2017densely}, so that the
parameter vectors $\theta_i \in \mathbb{R}^d$ of all clients are
dimensionally and structurally aligned; here $\theta_i$ denotes the
parameters of client $i$'s network. Before the first round, the server
distributes this common architecture together with its pretrained
initialization $\pmb{\theta}^{0}$ to all clients, so that training
starts from a shared pretrained backbone rather than from a random
initialization. Training proceeds over $T$ synchronous communication
rounds. At the beginning of round $t$, the server broadcasts the
current global model $\pmb{\theta}^{t}$ to all clients; each client $i$
initializes its local network from $\pmb{\theta}^{t}$ and performs $E$
epochs of local training on $D_i$, producing updated parameters
$\theta_i^{t}$. All clients participate in every round; no client
selection or filtering is applied. The server then forms the next
global model by weighted aggregation:
\begin{equation}
    \pmb{\theta}^{t+1} \leftarrow \sum_{i=1}^{K} \omega_i^{t} \, \theta_i^{t},
    \qquad \text{s.t.} \quad \sum_{i=1}^{K} \omega_i^{t} = 1, \;\; \omega_i^{t} \geq 0,
    \label{eq:aggregation}
\end{equation}
where $\omega_i^{t}$ is the aggregation weight assigned to client $i$
at round $t$.

In FedAvg~\citep{mcmahan2017communication}, the weights are fixed to
$\omega_i = n_i / N$ for all rounds. Under non-IID client data, this
static, purely size-proportional rule is known to degrade the quality
of the global model, and the resulting accuracy loss has been
attributed to the divergence of client weights, which grows with the
statistical distance between each client's label distribution and the
population distribution~\citep{zhao2018}. We attribute this limitation
to two implicit assumptions embedded in the rule, namely
(a)~that dataset size is a sufficient proxy for contribution quality,
and (b)~that the relative importance of clients is constant throughout
training.

Regarding~(a), weighting client updates purely by data volume treats
the merit of a client as the amount of data it holds, which has been
argued to be inappropriate in realistic deployments where clients
differ in the informativeness and not only in the size of their
data~\citep{zeng2021fedcav}. A client with a large but homogeneous
dataset may produce updates that are largely redundant with respect to
the global model, while a small client may hold underrepresented
patterns, for instance uncommon pathologies in a medical imaging
federation, whose contribution would be systematically undervalued by size-based weighting. Regarding~(b), client utility is not a fixed
property, since recent analysis of weighted aggregation shows that the
relative importance of a client to the global model changes over the
course of training, with a critical point in the learning dynamics
before which more coherent clients play the most essential role in
generalization~\citep{li2023revisiting}. In our setting, clients with
larger datasets tend to produce more stable, statistically
representative updates in early rounds, whereas as the global model
matures, every client, including those with small datasets, benefits
from the stronger initialization it receives, and a client that was
initially unreliable may subsequently produce updates that are well
aligned with the learning objective. Fixed weights are blind to this
evolution and permanently marginalize clients whose initial conditions
were unfavorable; adapting the weights to the per-round contribution of
each client has accordingly been shown to accelerate convergence under
non-IID data~\citep{wu2021fedadp}. Methods such as
FedProx~\citep{li2020fedprox} and
SCAFFOLD~\citep{karimireddy2020scaffold} address heterogeneity from a
different angle. FedProx modifies the local objective with a proximal
term, and SCAFFOLD corrects client drift using control variates.
Neither is designed to adapt the aggregation weights according to the
round-specific contribution of each client, and both therefore retain
the static weighting of FedAvg at the aggregation step.

A natural alternative is to replace dataset size with a measure of how
well each client's updated model aligns with the global model. L-DAWA,
for instance, aggregates client models according to the angular
divergence between each client's parameters and the global
model~\citep{rehman2023ldawa}, and FedSim guides aggregation using
similarity inferred from client
gradients~\citep{palihawadana2022fedsim}. However, similarity-based
weighting introduces a limitation of its own, which we refer to as a
cold-start effect. In the earliest rounds, the global model, even when
initialized from a pretrained backbone, is still task-agnostic and has
not yet incorporated meaningful knowledge from any client. At this
stage, high similarity between a client's model and the global model
does not necessarily represent a useful update, since it may merely
reflect that the client's data induced minimal parameter change,
leaving its model close to the uninformative initialization.
Conversely, low similarity may signal that the client's data contains
novel patterns the global model has not yet learned, which are
precisely the updates most needed for generalization. Trusting
similarity prematurely may therefore reward passivity and penalize
informativeness. This interpretation is our own design rationale rather
than an established result; it is, however, consistent with two
independent observations in the literature. First, using model
similarity directly as an aggregation weight has been reported to lack
sufficient adaptability and to introduce instability in the early
stages of training~\citep{shi2025fedawa}. Second, similarity-based
federated methods commonly precede the similarity computation by a
number of standard warm-up rounds, computing client similarities only
after the shared model has been trained
collaboratively~\citep{bdair2021fedperl}, an implicit acknowledgment
that similarity becomes meaningful only once the global model has
matured.

These complementary limitations, the staleness of fixed size-based
weights and the unreliability of premature similarity, motivate an
aggregation mechanism that (i)~uses dataset size as a stabilizing prior
in early rounds when similarity is unreliable, (ii)~progressively
transitions to similarity-based evaluation as the global model becomes
a credible reference, and (iii)~ensures that no client is permanently
excluded from meaningful participation. We realize all three objectives
through two coupled annealing mechanisms operating simultaneously, an
inner annealing governing the informational basis of client reputation,
and an outer annealing governing the distributional shape of the
resulting weights. We call their joint operation
\textbf{Fed}erated \textbf{D}ual \textbf{R}eputation \textbf{A}nnealing
\textbf{W}eighting (\textbf{FedDRAW}), which we describe in the
following subsections.

\subsection{Overview}
\label{sec:overview}

Figure~\ref{fig:feddraw_framework} gives an overview of the proposed
framework. We use the term \emph{dual annealing} for the joint
operation of two schedules that are annealed at the same time over the
communication rounds: (1) an inner annealing, which governs what
information defines the reputation of a client, and (2) an outer annealing,
which governs how strongly differences in reputation are turned into
differences in aggregation weight. At each round $t$, FedDRAW computes the aggregation weights $\omega_i^{t}$ through two operations performed on the server:

\begin{enumerate}

\item \textbf{Inner annealing (reputation scoring).} A reputation score
$\xi_i^{t}$ is computed for every client as a combination of its static
data-size prior and its dynamic similarity to the global model, with a
schedule that progressively shifts the reputation from the data-size
prior toward the similarity signal as training advances.

\item \textbf{Outer annealing (weight shaping).} The reputation scores
are converted into normalized aggregation weights through a
temperature-controlled softmax whose schedule progressively reduces
weight concentration, promoting an increasingly balanced and ultimately
equal influence of all participating clients on the global model, which
realizes the fairness objective of the method.

\end{enumerate}

Both schedules are governed by exponential functions of the round
index, in analogy with the cooling process of simulated
annealing~\citep{kirkpatrick1983simulated}, where a decreasing
temperature drives the transition from global exploration to local
refinement. In our setting, cooling captures the system's progression
from cautious, size-stabilized aggregation toward confident,
similarity-informed, and ultimately fair aggregation. The two stages
are not sequential phases, since they operate simultaneously at every
round, and their coupling is essential to the method
(Section~\ref{sec:coupling}). In the following subsections, we explain
the inner and the outer annealing in detail.

\vspace{6pt}

\begin{center}

\includegraphics[width=0.88\textwidth]{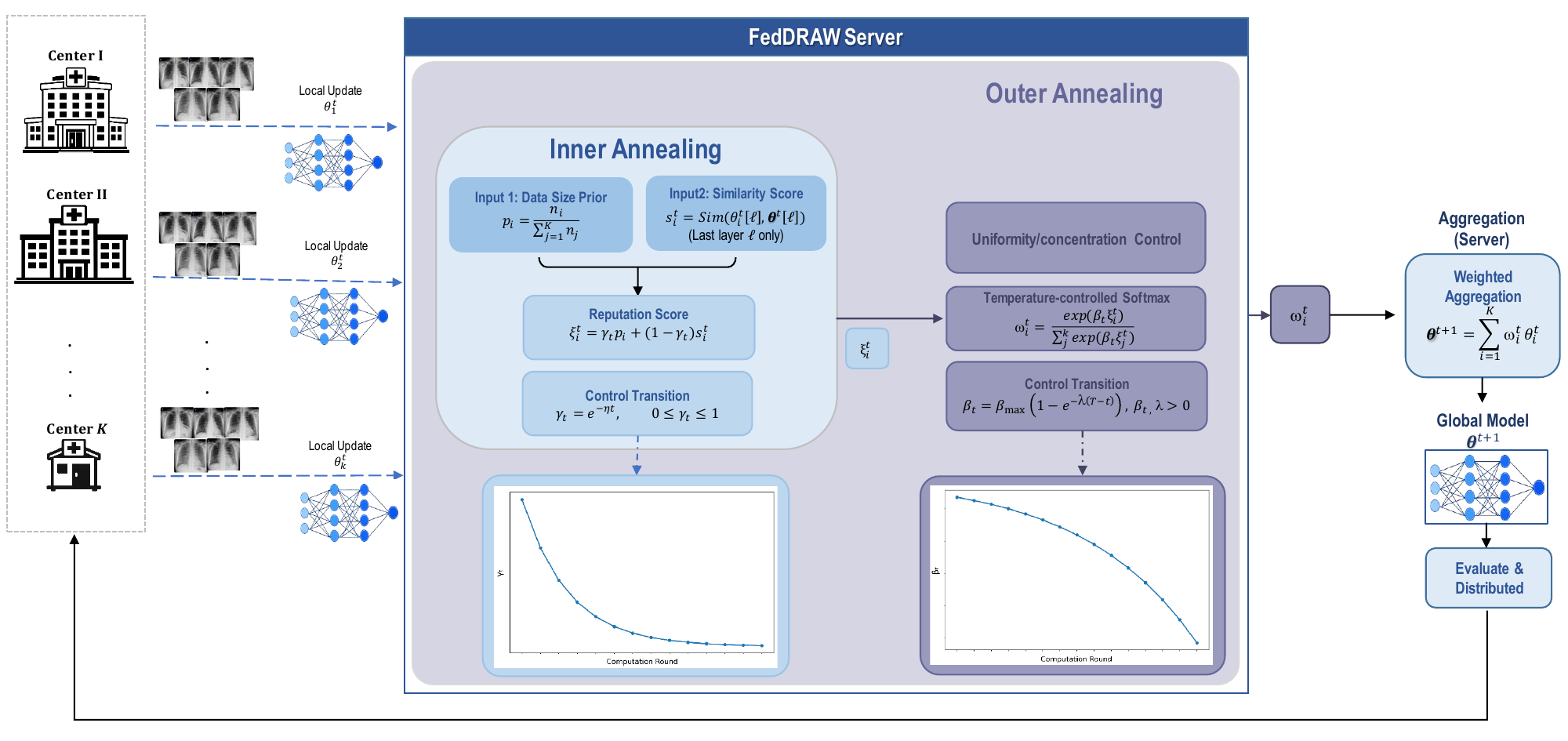}

\vspace{3pt}

\refstepcounter{figure}
\label{fig:feddraw_framework}

\noindent\parbox{\textwidth}{%
\scriptsize
\textbf{Figure \thefigure.} Overview of the FedDRAW framework. Clients perform
unmodified local training and upload their parameters $\theta_i^{t}$. On the
server, the inner annealing combines the static data-size prior $p_i$
with the last-layer similarity $s_i^{t}$ into a reputation score
$\xi_i^{t}$, and the outer annealing converts reputations into
aggregation weights $\omega_i^{t}$ through a temperature-controlled
softmax. The global model $\pmb{\theta}^{t+1}$ is then formed by
weighted aggregation and broadcast to all clients.
}

\end{center}

\vspace{6pt}

\subsection{Algorithm}
\label{sec:algorithm}

Algorithm~\ref{alg:feddraw} summarizes the complete FedDRAW procedure;
every step follows directly from the equations of
Sections~\ref{sec:inner}--\ref{sec:coupling}. Before walking through
it, we summarize the implementation characteristics of the method.
FedDRAW introduces no additional server-side model and no
client-to-server communication beyond the standard parameter upload of
any federated protocol, since all reputation and weight computations
(Eqs.~\eqref{eq:reputation}--\eqref{eq:beta_schedule}) are carried out
on the server from the parameters it receives in any case. Similarity
is computed on the parameters of the final classification layer only
(Section~\ref{sec:lastlayer}), since this layer carries the
task-specific representations most sensitive to each client's local
label distribution; restricting the computation to it also makes the
weighting step computationally efficient, costing
$\mathcal{O}(K \cdot d_\ell)$, where $d_\ell$ is the dimensionality of
the final layer, which is negligible relative to local training.
Because only the server-side weight computation is modified and the
client training procedure is left unchanged, FedDRAW is fully
compatible with client-side drift-mitigation methods (e.g.,
FedProx~\citep{li2020fedprox}) and serves as a drop-in replacement for
the FedAvg~\citep{mcmahan2017communication} aggregation step in any
standard federated pipeline.

\begin{algorithm}[t]
\caption{FedDRAW: Federated Dual Reputation Annealing Weighting}
\label{alg:feddraw}
\begin{algorithmic}[1]
\Require local datasets $\{D_i\}_{i=1}^{K}$; rounds $T$; local epochs $E$; inner annealing rate $\eta > 0$; outer annealing rate $\lambda > 0$; maximum inverse temperature $\beta_{\max} > 0$
\Ensure final global model $\pmb{\theta}^{T}$
\State $\pmb{\theta}^{0} \gets$ pretrained parameters of the DenseNet121 backbone, shared by all clients
\State $p_i \gets n_i \big/ \sum_{j} n_j$ for each client $i$
\Comment{Data Size Prior, Eq.~\eqref{eq:prior}}
\For{$t = 0, 1, \dots, T-1$}
    \Statex \hspace{\algorithmicindent}\textit{// Local training (in parallel on clients)}
    \For{each client $i = 1, \dots, K$ \textbf{in parallel}}
        \State $\theta_i^{t} \gets \operatorname{ClientUpdate}\!\left(D_i,\, \pmb{\theta}^{t},\, E\right)$
        \State send $\theta_i^{t}$ to the server
    \EndFor
    \Statex \hspace{\algorithmicindent}\textit{// Inner annealing (on server)}
    \State $\gamma_t \gets e^{-\eta t}$
    \Comment{Inner Annealing, Eq.~\eqref{eq:gamma}}
    \For{each client $i = 1, \dots, K$}
        \State $s_i^{t} \gets \mathrm{Sim}\!\left(\theta_i^{t}[\ell],\, \pmb{\theta}^{t}[\ell]\right)$
        \Comment{Similarity Score, Eq.~\eqref{eq:similarity}}
        \State $\xi_i^{t} \gets \gamma_t\, p_i + (1 - \gamma_t)\, s_i^{t}$
        \Comment{Reputation Score, Eq.~\eqref{eq:reputation}}
    \EndFor
    \Statex \hspace{\algorithmicindent}\textit{// Outer annealing (on server)}
    \State $\beta_t \gets \beta_{\max}\!\left(1 - e^{-\lambda\,(T-t)}\right)$
    \Comment{Outer Annealing, Eq.~\eqref{eq:beta_schedule}}
    \State $\displaystyle \omega_i^{t} \gets \exp\!\left(\beta_t\, \xi_i^{t}\right) \Big/ \sum_{j=1}^{K} \exp\!\left(\beta_t\, \xi_j^{t}\right)$ for each client $i$
    \Comment{emperature-Controlled Softmax, Eq.~\eqref{eq:softmax}}
    \Statex \hspace{\algorithmicindent}\textit{// Weighted aggregation (on server)}
    \State $\displaystyle \pmb{\theta}^{t+1} \gets \sum_{i=1}^{K} \omega_i^{t}\, \theta_i^{t}$
    \Comment{Weighted aggregation(global update), Eq.~\eqref{eq:server-update}}
    \State broadcast $\pmb{\theta}^{t+1}$ to all clients
\EndFor
\State \Return $\pmb{\theta}^{T}$
\end{algorithmic}
\end{algorithm}

Here $\operatorname{ClientUpdate}(D_i, \pmb{\theta}^{t}, E)$ denotes
$E$ epochs of local training on $D_i$ starting from the broadcast
global model $\pmb{\theta}^{t}$. All $K$ clients execute this update in
parallel at every round; the size priors $p_i$ are computed once from
$D_i$ before training and never recomputed.

\subsection{Inner Annealing: Reputation Scoring}
\label{sec:inner}

The central idea of the inner annealing stage is a gradual shift of the
weighting criterion. Aggregation initially relies on client data size,
exactly as in FedAvg~\citep{mcmahan2017communication}, where the weight
of client $i$ is proportional to its sample count $n_i$. It then
progressively transitions toward relying on the similarity between each
client's updated model and the global model. The stage thus answers the
question of what information should determine client importance at each
point in training. Its design principle is that the reputation score should rest on the dataset size while the global model is still immature, and should incorporate the client-to-global similarity only once that reference has stabilized. We state this as a design principle of FedDRAW rather than as
an established result; the practice of postponing similarity-based
reasoning until after a warm-up phase~\citep{bdair2021fedperl} and the
finding that the relative importance of clients changes over the course
of training~\citep{li2023revisiting} both point in the same direction.

To see why this direction of transition is appropriate, consider the
alternatives. Using only dataset size throughout reduces to
FedAvg~\citep{mcmahan2017communication}, with all of its limitations,
namely static weights that cannot adapt to evolving client quality, and
permanent marginalization of small but informative clients. Using only
similarity throughout lets the cold-start effect dominate the early
rounds, because a client whose data induces small parameter changes
appears highly similar to the immature global model not because it is
well aligned with the objective, but because it barely moved from the
starting point, while a client with diverse or complex data produces
large updates that lower its similarity despite being exactly what the
global model needs. The proposed schedule therefore begins with the size prior for stability and shifts, in an annealing-inspired manner, toward similarity once the global model provides a stable reference for the comparison.

\subsubsection{Data Size Prior}
For each client $i$, we define the normalized size prior
\begin{equation}
    p_i = \frac{n_i}{\sum_{j=1}^{K} n_j},
    \label{eq:prior}
\end{equation}
computed once before training and held constant. It encodes the
assumption that, absent any other information, a client with more data
is expected to produce a more representative update.

\subsubsection{Similarity Score}
At each round $t$, after receiving the updated parameters
$\theta_i^{t}$, the server computes
\begin{equation}
    s_i^{t}
    =
    \mathrm{Sim}\!\left(
        \theta_i^{t},\,
        \pmb{\theta}^{t}
    \right),
    \label{eq:similarity}
\end{equation}
where $\mathrm{Sim}(\cdot,\cdot)$ denotes cosine similarity applied to
model parameters.

The restriction of this computation to the parameters of the final
layer is an implementation choice orthogonal to the annealing
framework; we motivate it in Section~\ref{sec:lastlayer}. Unlike the
size prior, $s_i^{t}$ is fundamentally dynamic, since it changes at
every round because both $\theta_i^{t}$ and $\pmb{\theta}^{t}$ evolve,
reflecting the current state of learning rather than a static property
of the data.

\subsubsection{Reputation Score}
We define the reputation score of client $i$ at round $t$ as the convex
linear combination
\begin{equation}
    \xi_i^{t} = \gamma_t\, p_i + (1 - \gamma_t)\, s_i^{t},
    \label{eq:reputation}
\end{equation}
where $\gamma_t \in [0,1]$ is the inner annealing parameter, which
controls the transition within the inner annealing, in that it
determines how much the combination relies on the data-size prior
versus the similarity signal at each round. At $\gamma_t = 1$ the
reputation is the pure size prior; at $\gamma_t = 0$ it is the pure
similarity score; intermediate values blend the two. Our design
requires $\gamma_t$ to decrease from $1$ toward $0$ over the course of
training, encoding the progressive shift of trust from the static prior
to the dynamic similarity signal.

\subsubsection{Control Transition: Exponential Decay Schedule for $\gamma_t$}
\label{sec:gamma-schedule}
We require a schedule with three properties: (i)~$\gamma_0 = 1$,
ensuring pure size-based reputation at the first round;
(ii)~$\gamma_t \to 0$ as $t \to \infty$, ensuring eventual reliance on
similarity; (iii)~smooth decay without abrupt transitions that could
destabilize training. Inspired by the exponential cooling schedule of
simulated annealing~\citep{kirkpatrick1983simulated,laarhoven1987}, we
adopt
\begin{equation}
    \gamma_t = \exp(-\eta\, t),
    \label{eq:gamma}
\end{equation}
where $\eta > 0$ is the inner annealing rate. The exponential form of
Eq.~\eqref{eq:gamma} shapes the transition in a specific way, since the
decrease of $\gamma_t$ is fastest in the first rounds and becomes
progressively slighter thereafter, so that the bulk of the shift away
from the size prior happens early, followed by a long, gentle approach
toward pure similarity-based reputation. The analogy to simulated
annealing is more than superficial. At high temperature, that is, for
large $\gamma_t$ in the early rounds, the aggregation draws broadly
from all clients weighted by data size, a form of exploration that does
not prejudge any client based on similarity to an immature model, and
as $\gamma_t$ cools, the aggregation exploits the similarity signal to
focus on clients aligned with the now-mature global model.

\paragraph{Role of $\eta$.}
The rate $\eta$ governs the speed of the transition and should be
calibrated to the convergence dynamics of the learning task. A large
$\eta$ yields rapid decay, appropriate when the global model converges quickly, for instance when training starts from a strong pretrained backbone, and a stable reference for the similarity computation is therefore reached early. A small $\eta$
maintains the stabilizing size prior for many rounds, appropriate for
highly heterogeneous settings where the global model needs longer to
become a meaningful reference. More generally, the suitable range of
$\eta$ depends on the total number of communication rounds of the
federation and on the characteristics and quality of the client data,
and should be chosen with both in mind. We used a single intermediate
value throughout all experiments without scenario-specific tuning; the
exact value is reported in the experimental setup.

\subsection{Outer Annealing: From Selective to Fair Aggregation}
\label{sec:outer}

The inner stage determines the informational content of client
reputation; a separate concern is how reputation scores translate into
weights, that is, the distributional shape of the weight vector. Even
perfectly accurate reputations, if converted into a distribution that
assigns negligible weight to most clients, raise a fundamental fairness
concern.

Client-level fairness has been formalized in federated learning in
several ways, for instance through minimax objectives over mixtures of
client distributions~\citep{mohri2019agnostic} or through the
uniformity of model performance across devices~\citep{li2021ditto}. Our
concern here is a distinct, aggregation-level notion, which we
operationalize as the requirement that no client should be permanently
excluded from meaningful participation in the global model, that is,
that no client should be assigned a persistently negligible aggregation
weight. Ethically, federated collaborations are formed under the
expectation that participation benefits all parties; an aggregation
scheme that systematically ignores certain clients undermines that
cooperative foundation. Technically, permanent exclusion creates a
self-reinforcing negative cycle, in which a client receiving near-zero
weight exerts no influence on $\pmb{\theta}^{t+1}$, so the next global
model is poorly aligned with its local data, producing larger local
deviations, lower similarity scores, and still lower weights. This
compounding marginalization is especially damaging because marginalized
clients often hold precisely the rare or underrepresented patterns
(e.g., uncommon diagnoses) that the global model needs to generalize.

The outer stage controls how strongly reputation differences translate
into weight differences, and ensures this strength decreases over time.
Early in training, a peaked distribution favoring high-reputation
clients is desirable, because the inner stage is supplying size-based
reputations, and amplifying size differences helps the global model
stabilize by attending to the largest, most representative datasets.
Late in training, a near-uniform distribution is desirable, because all
clients have by then benefited from many rounds of global updates and
improved their local models, and the initially weak or dissimilar among
them should no longer be marginalized. The transition between the two
regimes should be gradual, avoiding abrupt weight redistribution that
could destabilize the global model.

\subsubsection{Temperature-Controlled Softmax}
We convert the reputation scores obtained from the inner annealing
(Eq.~\eqref{eq:reputation}) into normalized weights through a softmax
with time-dependent inverse temperature $\beta_t > 0$, that is,
$\beta_t = 1/T_t$ in the temperature notation of Hinton et
al.~\citep{hinton2015}:
\begin{equation}
    \omega_i^{t} = \frac{\exp\!\left(\beta_t\, \xi_i^{t}\right)}{\sum_{j=1}^{K} \exp\!\left(\beta_t\, \xi_j^{t}\right)}.
    \label{eq:softmax}
\end{equation}
Here, $\omega_i^{t}$ is the final aggregation weight of client $i$. In
the early and middle rounds, it is driven by the reputation produced by
the inner annealing, that is, the blend of data size and similarity,
whereas in the late rounds the outer schedule takes over and drives all
weights toward equality. The parameter $\beta_t$ is thus responsible
for the shift from reputation-driven to equal weighting, which is
precisely the notion of fairness realized in our method. Large
$\beta_t$ amplifies even small reputation differences; in the limit
$\beta_t \to \infty$, the entire weight concentrates on the single
highest-reputation client. Small $\beta_t$ compresses differences; in
the limit $\beta_t \to 0$, the weights become uniform,
$\omega_i^{t} \to 1/K$. This makes $\beta_t$ the natural control
parameter for the peaked-to-uniform transition we seek.

\subsubsection{Control Transition: Deferred Annealing Schedule for $\beta_t$}
\label{sec:beta-schedule}

The outer temperature $\beta_t$ controls how sharply the softmax
converts reputation scores into aggregation weights. The desired behavior is that the weights remain selective through the early and middle rounds and equalize only toward the end of training, since premature equalization risks discarding the contribution of small
but informative clients before it has been absorbed into the global
model. We therefore employ the schedule
\begin{equation}
\beta_t \;=\; \beta_{\max}\!\left(1-e^{-\lambda\,(T-t)}\right),
\qquad t=1,\dots,T,
\label{eq:beta_schedule}
\end{equation}
where $\beta_{\max}>0$ denotes the maximum inverse temperature,
$\lambda>0$ the outer annealing rate, and $T$ the total number of
communication rounds. The schedule is parameterized by the number of
remaining rounds $T-t$ rather than by elapsed time. Consequently,
$\beta_t$ remains close to $\beta_{\max}$ throughout the early and
middle phases of training, preserving selective aggregation, and decreases only as $t$
approaches $T$. At the final round, the exponent of the softmax
vanishes, so that $\beta_T=0$ and the aggregation weights become
exactly uniform, $\omega_i^{T}=1/K$. Differentiated weighting in the
early and middle rounds and equal weighting at convergence are thus
obtained by construction, independently of the values of $\beta_{\max}$
and $\lambda$. Because the flattening of the weight distribution is
postponed to the final rounds rather than beginning immediately, we
refer to Eq.~\eqref{eq:beta_schedule} as a deferred annealing schedule.

\paragraph{Role of $\beta_t$.}
At each round, $\beta_t$ acts as the inverse temperature of the softmax
and thus determines how strongly differences in reputation are
translated into differences in aggregation weight, since for large
$\beta_t$ the weights concentrate on high-reputation clients, whereas
$\beta_t\to0$ yields the uniform distribution $1/K$ regardless of the
reputation scores. The magnitude of this selectivity is governed by
$\beta_{\max}$, since a larger value concentrates aggregation weight
more heavily on high-reputation clients, whereas a smaller value keeps
the distribution closer to uniform throughout. The rate $\lambda$ sets
how gradually the flattening unfolds over the final rounds, since a
larger $\lambda$ keeps $\beta_t$ near $\beta_{\max}$ until immediately
before the last round, whereas a smaller $\lambda$ begins the descent
earlier. Both parameters affect only the magnitude and timing of the
deviation from uniform weighting, not the terminal behavior; the values
used in our experiments are reported in the experimental setup.

\paragraph{Interaction with the inner schedule.}
The two annealing schedules operate on complementary time scales. The
inner schedule $\gamma_t=e^{-\eta t}$ shifts the reputation scores from
the data-size prior toward the similarity signal during the early
rounds, whereas the outer schedule in Eq.~\eqref{eq:beta_schedule}
flattens the weight distribution only in the final rounds. The requirement that the transition from the size prior to the similarity signal completes before the weights equalize therefore holds structurally for any $\eta>0$ and $\lambda>0$,
and no coupling constraint of the form $\lambda<\eta$ needs to be
imposed between the two schedules.

\subsection{Weighted Aggregation}
\label{sec:coupling}

The inner and outer stages operate simultaneously at every round,
forming a coupled dual annealing process. At round $t$: (1)~the inner
stage computes $\gamma_t$ and the reputation vector
$\{\xi_i^{t}\}_{i=1}^{K}$ via Eq.~\eqref{eq:reputation}; (2)~the outer
stage computes $\beta_t$ and the weight vector
$\{\omega_i^{t}\}_{i=1}^{K}$ via Eq.~\eqref{eq:softmax}. The two
parameters decay on their own schedules but govern complementary
aspects of aggregation, in that $\gamma_t$ controls the content of
client evaluation, that is, what defines reputation, while $\beta_t$
controls its consequence, that is, how sharply reputation differences
become weight differences.

An analogy clarifies the interaction. The inner annealing acts as a
scorer evaluating each client's contribution quality, while the outer
annealing acts as a supervisor deciding how strongly to act on those
scores. Early on, the scorer relies on data size, and the supervisor
amplifies the resulting differences, yielding stable, size-driven,
selective aggregation. As training advances, the scorer shifts toward
similarity while the supervisor simultaneously relaxes toward equal
treatment of all clients. The combined trajectory moves from informed
and selective aggregation to informed and fair aggregation, realizing a
stabilize-first, equalize-later principle.

The coupling is essential. Outer annealing without the inner stage
would still equalize weights over time, but early-round reputation
would rest on a single static signal, missing the opportunity to
leverage the right information at the right time. Inner annealing
without the outer stage would transition to similarity-based reputation
but could permanently suppress low-similarity clients even at late
rounds, violating fairness. Only the joint operation achieves all three
design goals, namely early stability, a gradual mid-training change of criterion, and late-training fairness.

After computing the weights $\{\omega_i^{t}\}$ through dual annealing,
the server forms
\begin{equation}
    \pmb{\theta}^{t+1} \leftarrow \sum_{i=1}^{K} \omega_i^{t} \, \theta_i^{t}.
    \label{eq:server-update}
\end{equation}
This aggregation step is identical in form to standard federated
averaging; the entirety of FedDRAW's novelty lies in the dynamic,
round-dependent computation of $\omega_i^{t}$.

\subsection{Last-Layer Similarity}
\label{sec:lastlayer}

We compute cosine similarity using only the parameters of the final
(classification) layer rather than the full parameter vector. This
choice is grounded in what different layers of a deep network actually
represent, and is particularly consequential in federated settings.

Representations learned by deep networks progress from generic to
task-specific with depth, since early layers encode features such as
edges and colors that are applicable to many datasets and tasks, and
the representation must transition from general to specific by the
final layers of the network~\citep{yosinski2014}. When all clients
start each round from the same global model $\pmb{\theta}^{t}$ and
perform only a small number of local epochs $E$, the generic
early-layer parameters therefore remain both close to their round-start
values and highly similar across clients, since they encode features
that are largely independent of the particular local dataset. When
similarity is computed over the full parameter vector
$\theta_i^{t} \in \mathbb{R}^d$, this large mass of nearly identical
generic parameters dominates the score and drives it toward
artificially high values, so that the discriminative signal of
client-specific learning is drowned out. Consistent with this, using
whole-model similarity directly has been reported to insufficiently
capture the relationship between clients~\citep{shi2025fedawa}.

The final layer, in contrast, maps learned feature representations to
output predictions and is the component most directly tied to each
client's local label distribution, and it has been shown to be the part
of the network most strongly biased by label-distribution skew across
clients~\citep{luo2021nofear}. This has also been exploited directly in
federated settings, where the weights of the last layer of a local
update are more sensitive to the local data distribution than those of
other layers, which is why the cosine similarity of the last layer
between the global model and each client update carries a usable signal
about that client's data~\citep{yaldiz2023cosdefense}. Computing
similarity on this layer therefore yields a sharper, more
discriminative measure of client-specific learning dynamics, and does
so at negligible cost. We additionally verified this behavior directly.
Two models pre-trained on ImageNet~\citep{deng2009} and fine-tuned on
disjoint 5-class datasets (animals versus cars) yield a full-network
cosine similarity close to $1.0$, misleadingly high and dominated by
the shared pre-trained parameters, whereas restricting the computation
to the last layer drops the score to approximately $0.2$, correctly
reflecting the substantial divergence in what the two models learned
during fine-tuning.

%=====================================================================
% Results section
%=====================================================================
\section{Experimental setup and evaluation}
\label{sec:results}

In this section, we evaluate the proposed method and, in particular,
the interaction between its inner and outer annealing schedules, which
jointly control how the aggregation weights evolve from
size-prior-dominated to similarity-informed and finally to near-uniform
values over the course of federated training. FedDRAW is compared
against seven well-established federated learning algorithms,
introduced in Section~\ref{sec:related}: FedAvg
\citep{mcmahan2017communication}, FedNova \citep{wang2020fednova},
FedAdp \citep{wu2021fedadp}, FedProx \citep{li2020fedprox}, SCAFFOLD
\citep{karimireddy2020scaffold}, MOON \citep{li2021moon}, and FedDyn
\citep{acar2021feddyn}. The evaluation is carried out on two public
multi-label chest radiograph datasets, CheXpert
\citep{irvin2019chexpert} and ChestMNIST \citep{yang2023medmnist},
under twelve client-partition scenarios of varying difficulty.

Because all methods were trained with identical local hyperparameters
(Section~\ref{sec:experimental_setup}), the comparison isolates the
effect of the respective heterogeneity-handling strategy. We first
describe the experimental setup, the evaluation metrics, and the design
of the partition scenarios, and then present the results on each
dataset, a statistical comparison across all scenarios, and an analysis
of the FedDRAW aggregation dynamics.

% ---------------------------------------------------------------------
\subsection{Experimental setup}
\label{sec:experimental_setup}

All experiments were conducted with a DenseNet121 backbone initialized
with weights pretrained on ImageNet, in which the final classification
layer was adapted to the five target pathologies (Atelectasis, Edema,
Pleural Effusion, Cardiomegaly, and Consolidation). We replaced the
softmax cross-entropy loss of DenseNet121 with a multi-label binary
cross-entropy loss with sigmoid activation and trained the resulting
model on the clients. Uncertain and missing CheXpert labels were
converted to binary values using the U-Ones/U-Zeros mapping described
in Section~\ref{sec:datasets}. Label smoothing with a factor of $0.05$
and gradient-norm clipping at $5.0$ were applied to stabilize local
training, and a weight decay of $5\times10^{-3}$ was used.

All images were converted to three-channel grayscale, resized to the
network input resolution, and enhanced with contrast-limited adaptive
histogram equalization (CLAHE; kernel size $28$, clip limit $0.02$)
before being normalized with the channel mean and standard deviation of
the training data ($0.5146$ and $0.2548$, respectively). This
preprocessing is deterministic and was applied identically to the
training, validation, and test images. During local training, the
images were additionally augmented geometrically with a random rotation
of up to $\pm 8^\circ$ and a random affine transformation with
translations of up to $9\%$ of the image size and isotropic scaling in
$[0.95, 1.05]$, applied stochastically. Validation and test images
received no augmentation. The same pipeline was used at every client,
and no artificial domain shift was introduced between clients, so that
the heterogeneity in each scenario arose exclusively from the
partitioning of the samples and their label distributions.

For CheXpert, the 191,027 frontal training radiographs were partitioned
among the federated clients, while the frontal subsets of the validation
and test partitions provided with the dataset (202 of 234 and 518 of 668
images, respectively) were used for threshold selection and final
evaluation~\citep{irvin2019chexpert}. For ChestMNIST, we used the
official MedMNIST split of 78,468 training, 11,219 validation, and
22,433 test images at a resolution of $224\times224$ pixels from the
large-size release (MedMNIST+)~\citep{yang2023medmnist}; only the
training partition was distributed among the clients.

In each communication round, every client performed $E = 2$ local
epochs of training with the Adam optimizer, a learning rate of
$1\times10^{-5}$, and a batch size of 64, and federated training was
run for $T = 15$ communication rounds. Identical training
hyperparameters were used for FedDRAW and for all competing methods, so
that the reported differences are attributable to the aggregation
strategies rather than to differences in local optimization. Each
experiment was repeated with five independent runs, and we report the
mean and standard deviation across runs. The FedDRAW-specific
hyperparameters were kept fixed across all scenarios and both datasets,
CheXpert and ChestMNIST ($\eta = 0.4$, $\lambda = 0.2$,
$\beta_{\max} = 3$).

% ---------------------------------------------------------------------
\subsection{Evaluation metrics}
\label{sec:evaluation_metrics}

Model performance was evaluated using two complementary metrics, namely
AUC and GM, that is, the square root of the product of sensitivity and
specificity, both macro-averaged over the five target pathologies. AUC
is the standard metric in chest radiograph classification, and prior
studies applying federated learning to chest X-ray datasets such as
CheXpert report their evaluation primarily in terms of
AUC~\citep{ziegler2022defending,zulqarnain2026federated,arasteh2023enhancing}.
AUC alone, however, does not reveal the consequences of severe class
imbalance, and the same holds for accuracy, which is dominated by the
prevalent negative class, so that a classifier which predicts every
finding as absent still attains a high value. AUC conceals the
imbalance for a different and less obvious reason. The AUC is
constructed from the true-positive and false-positive rates, each of
which is normalized by the size of its own class, so the curve, and
hence the area beneath it, is invariant to the ratio of positive to
negative samples. A model can therefore rank the positive cases well
overall, and be credited with a high AUC, while at every threshold that
could actually be deployed it detects only a small fraction of the
positive cases. Because AUC additionally aggregates over all possible
thresholds, it also does not describe the behavior of the classifier at
the single threshold at which it is operated. This limitation is
particularly important for chest X-ray datasets, whose labels are both
highly imbalanced and affected by label noise arising from automated
report labeling; under such conditions, evaluations restricted to
prevalence-invariant and threshold-independent metrics can give an
incomplete or misleading picture of clinical
usefulness~\citep{cabitza2026why}.
We therefore additionally report GM~\citep{kubat1997addressing},
defined for a pathology $c$ as
\begin{equation}
\mathrm{GM}_c \;=\; \sqrt{\mathrm{Se}_c \cdot \mathrm{Sp}_c},
\qquad
\mathrm{Se}_c = \frac{\mathrm{TP}_c}{\mathrm{TP}_c + \mathrm{FN}_c},
\qquad
\mathrm{Sp}_c = \frac{\mathrm{TN}_c}{\mathrm{TN}_c + \mathrm{FP}_c},
\label{eq:gm}
\end{equation}
where $\mathrm{Se}_c$ and $\mathrm{Sp}_c$ denote the sensitivity and
specificity of pathology $c$ at its decision threshold, and the
reported value is the macro-average of $\mathrm{GM}_c$ over the five
target pathologies. Because the GM collapses toward zero whenever
either sensitivity or specificity degenerates, GM cannot be inflated by
biasing predictions toward the majority class, which makes it a
necessary complement to AUC on strongly imbalanced data. Importantly, a
model can attain a high AUC in a given round while simultaneously
exhibiting a low GM, because AUC summarizes the ranking of predicted
probabilities over all possible thresholds, whereas GM evaluates the
single threshold at which the classifier is actually operated.
Evaluating both therefore reveals operating-point behavior that AUC
alone cannot capture.

Any threshold-dependent metric requires an explicit decision rule, and
the default threshold of $0.5$ is not appropriate for chest radiograph
classification. That value implicitly assumes balanced classes and
calibrated probabilities, neither of which holds here. With positive
prevalences well below $50\%$ for several of the target pathologies, a
fixed threshold of $0.5$ systematically biases predictions toward the
negative class, suppressing sensitivity while leaving specificity
near-saturated, so that the resulting GM reflects the prevalence of the
data rather than the discriminative quality of the model. The threshold
must therefore be determined from data, and it must be determined per
pathology, since the five findings differ substantially in prevalence
and in the sharpness of their probability distributions.

Inspired by the threshold-selection protocol used for chest radiograph
classification by Tiu et al.~\citep{tiu2022expert}, we determine the
decision thresholds on the validation set and evaluate on the test set.
For every algorithm and every run, the per-pathology thresholds are
selected at the final communication round subject to maximizing
$\mathrm{GM}_c$ on the validation set, and are then applied once,
without modification, to the held-out test set.

% ---------------------------------------------------------------------
\subsection{Client partition scenarios}
\label{sec:scenarios}

The two datasets alone do not cover the range of client configurations
under which the objective of FedDRAW becomes testable, since neither
provides a predefined partition into institutions with differing sizes
or differing pathology distributions. It is therefore necessary to
construct federated scenarios artificially, so that the behavior of
FedDRAW and of the competing methods can be compared under controlled
and reproducible degrees of heterogeneity. The scenarios were designed
to test the central hypothesis of this work, namely that a client with
few local samples may nevertheless be highly informative, either
because its data quality is high relative to its size or because it
holds samples of a pathology that is scarce or absent at the larger
clients, and that such a client should not be marginalized by
sample-size-proportional aggregation. In a centralized reference evaluation, models trained on the data of the designated small clients achieved higher performance than models trained on the data of the individual larger clients. This gap was built into the partition design and describes the data rather than the behavior of the aggregation method. For each dataset, we
constructed a dedicated set of scenarios.
Tables~\ref{tab:scenarios_chx} and~\ref{tab:scenarios_chm} give the
client-level sample counts. We denote
the CheXpert scenarios CHX-1 to CHX-7 and the ChestMNIST scenarios
CHM-1 to CHM-5.

The CheXpert scenarios, listed in Table~\ref{tab:scenarios_chx}, follow
a deliberate progression. CHX-1 to CHX-4 vary the ratio of large to
small clients in order to test whether FedDRAW recognizes the valuable
information of the smaller clients, rather than letting the larger
clients dominate the aggregation, across increasingly asymmetric
federations. CHX-4 is the most severe of these configurations, as the
small client contributes only $2.7\%$ of the total training data.
CHX-5 and CHX-6 simulate the clinically relevant situation in which a
small client holds informative samples of a specific pathology that the
larger clients largely lack. The small client of CHX-5 was enriched
with 3{,}500 Atelectasis-positive samples and the small client of CHX-6
with 3{,}500 Edema-positive samples, in both cases alongside samples of
the remaining four pathologies. In these two scenarios, the small client
holds $20\%$ of the training data but approximately $65\%$ of all
positive samples of the enriched pathology, whereas each large client
contains only 950 of them. Finally, in CHX-7, the CheXpert training set
was split into four clients uniformly at random; this homogeneous
reference scenario verifies that FedDRAW does not degrade performance
when no meaningful heterogeneity is present and can compete closely
with the other methods.

% ============================================================
% TABLE 2 -- CheXpert
% ============================================================
\begin{center}

\refstepcounter{table}
\label{tab:scenarios_chx}

\noindent\parbox{\textwidth}{%
\scriptsize
\textbf{Table \thetable.} Client sample sizes in the CheXpert scenarios.
C$i$ denotes the $i$-th federated client. Each scenario represents a
different heterogeneity setting: CHX-1 to CHX-4 vary the ratio of large
to small clients, CHX-5 and CHX-6 enrich the small client with informative
samples of a specific pathology (Atelectasis and Edema, respectively),
and CHX-7 is a homogeneous random split. In all heterogeneous scenarios,
the small clients carry a disproportionate share of informative samples
relative to their size.
}

\vspace{4pt}

\footnotesize
\setlength{\tabcolsep}{4pt}
\renewcommand{\arraystretch}{1.0}

\begin{tabular}{llrr}
\toprule
Scenario & Composition & Samples per client & Total \\
\midrule
CHX-1 & 1 large + 3 small & C1: 20{,}000; C2--C4: 5{,}000 each & 35{,}000 \\
CHX-2 & 2 large + 2 small & C1--C2: 20{,}000 each; C3--C4: 5{,}000 each & 50{,}000 \\
CHX-3 & 3 large + 1 small & C1--C3: 10{,}000 each; C4: 2{,}500 & 32{,}500 \\
CHX-4 & 9 large + 1 small & C1--C9: 10{,}000 each; C10: 2{,}500 & 92{,}500 \\
CHX-5 & Atelectasis-enriched small client & C1: 5{,}000; C2--C3: 10{,}000 each & 25{,}000 \\
CHX-6 & Edema-enriched small client & C1: 5{,}000; C2--C3: 10{,}000 each & 25{,}000 \\
CHX-7 & 4 equal (random split) & C1: 47{,}756; C2--C4: 47{,}757 each & 191{,}027 \\
\bottomrule
\end{tabular}

\end{center}
% ============================================================
% TABLE 3 -- ChestMNIST
% ============================================================
\begin{center}

\refstepcounter{table}
\label{tab:scenarios_chm}

\noindent\parbox{\textwidth}{%
\scriptsize
\textbf{Table \thetable.} Client sample sizes in the ChestMNIST scenarios.
C$i$ denotes the $i$-th federated client. CHM-1 to CHM-3 create increasingly
severe quantity imbalance; CHM-4 enriches the small client with informative
Atelectasis samples, and CHM-5 is a homogeneous reference split. In all
heterogeneous scenarios, the small clients carry a disproportionate share
of informative samples relative to their size.
}

\vspace{4pt}

\footnotesize
\setlength{\tabcolsep}{4pt}
\renewcommand{\arraystretch}{1.0}

\begin{tabular}{llrr}
\toprule
Scenario & Composition & Samples per client & Total \\
\midrule
CHM-1 & 1 large + 1 small & C1: 18{,}000; C2: 60{,}468 & 78{,}468 \\
CHM-2 & 1 large + 3 small & C1--C3: 9{,}500 each; C4: 49{,}968 & 78{,}468 \\
CHM-3 & 5 large + 1 small & C1--C5: 14{,}000 each; C6: 8{,}468 & 78{,}468 \\
CHM-4 & Atelectasis-enriched small client & C1: 7{,}000; C2--C4: 15{,}000 each & 52{,}000 \\
CHM-5 & 4 equal & C1--C4: 19{,}617 each & 78{,}468 \\
\bottomrule
\end{tabular}

\end{center}

The ChestMNIST scenarios, listed in Table~\ref{tab:scenarios_chm},
mirror this design on a smaller and noisier dataset. CHM-1 and CHM-2
create pronounced quantity imbalance with one and three small clients,
respectively, and CHM-3 extends the imbalance to a six-client
federation in which the small client holds only $10.8\%$ of the data,
making it the most severe ChestMNIST configuration. CHM-4 is the
class-enriched counterpart of CHX-5, with the small client enriched for
Atelectasis (4{,}000 positive samples, i.e., more than half of its local
data and the majority of all Atelectasis-positive samples in the
federation), and CHM-5 is the homogeneous
four-client reference split.

\subsection{Results on CheXpert}
\label{sec:results_chexpert}

Table~\ref{tab:results_chx} summarizes the performance of FedDRAW and
the competing methods on CheXpert across the seven scenarios. FedDRAW
attained the highest GM in every evaluated configuration and the
highest AUC in five of the seven; in the remaining two scenarios its
AUC was within 0.19 and 0.16 percentage points of the best value. The
magnitude of the advantage varied systematically with the degree and
type of heterogeneity, and in the quantity-skewed scenarios the
improvements in GM consistently exceeded those in AUC. This pattern is
directly aligned with the central design objective of FedDRAW. Under
sample-size-proportional aggregation, the updates of a small client are
strongly diluted regardless of their quality, whereas FedDRAW can temporarily amplify such updates when their similarity to the current global model is high and
subsequently anneals the weights back toward uniformity. Since the
reallocated influence primarily changes where the model operates rather
than how it ranks, the benefit concentrates in the threshold-dependent
metric.
CHX-4 is the most severe quantity-skewed configuration, in which the
small client holds only $2.7\%$ of the training data and faces nine
large clients that could easily dominate the aggregation. Even under
this extreme asymmetry, FedDRAW remains reliable. It attained the
highest GM of $77.35 \pm 0.13\%$, improving over FedAvg
($75.61 \pm 0.27\%$) by 1.74 percentage points, while its AUC of
$85.57 \pm 0.60\%$ lay within 0.19 percentage points of the best value
in this scenario, a difference well inside one standard deviation. The
two metrics therefore do not agree on the ordering of the methods, and
the disagreement is informative rather than incidental. Because AUC is
invariant to prevalence and averages over thresholds that are never
applied, it cannot register the effect that the reallocated influence
has on the model, since the small client shifts how positive and
negative cases are separated at the selected operating points, which is
exactly what GM measures and what AUC is blind to. Under the metric
that reflects the deployed decision rule, FedDRAW is the leading method
even in the most asymmetric federation considered here.

\begin{table}[!htbp]
\centering
\refstepcounter{table}
\label{tab:results_chx}
\noindent\parbox{\linewidth}{%
\scriptsize
\textbf{Table \thetable.} Performance (\%) on CheXpert across the seven
client-partition scenarios (mean $\pm$ standard deviation over five runs).
Best values per column are shown in bold.
}
\vspace{4pt}
\resizebox{\textwidth}{!}{%
\begin{tabular}{lcccccccc}
\toprule
& \multicolumn{2}{c}{CHX-1} & \multicolumn{2}{c}{CHX-2}
& \multicolumn{2}{c}{CHX-3} & \multicolumn{2}{c}{CHX-4} \\
\cmidrule(lr){2-3}\cmidrule(lr){4-5}\cmidrule(lr){6-7}\cmidrule(lr){8-9}
Algorithm & AUC & GM & AUC & GM & AUC & GM & AUC & GM \\
\midrule
FedDRAW  & \textbf{86.67 $\pm$ 0.25} & \textbf{78.76 $\pm$ 0.18} & \textbf{87.46 $\pm$ 0.30} & \textbf{77.83 $\pm$ 0.81} & \textbf{86.64 $\pm$ 0.18} & \textbf{78.03 $\pm$ 0.91} & 85.57 $\pm$ 0.60 & \textbf{77.35 $\pm$ 0.13} \\
FedAvg   & 84.98 $\pm$ 0.68 & 76.01 $\pm$ 1.31 & 85.77 $\pm$ 0.33 & 75.41 $\pm$ 1.32 & 85.71 $\pm$ 0.43 & 76.71 $\pm$ 1.80 & \textbf{85.76 $\pm$ 0.68} & 75.61 $\pm$ 0.27 \\
FedNova  & 85.33 $\pm$ 0.75 & 76.68 $\pm$ 0.35 & 86.16 $\pm$ 0.24 & 76.40 $\pm$ 1.00 & 85.63 $\pm$ 0.78 & 77.03 $\pm$ 0.39 & 82.88 $\pm$ 3.15 & 76.20 $\pm$ 0.27 \\
FedProx  & 85.40 $\pm$ 0.33 & 76.26 $\pm$ 1.46 & 86.00 $\pm$ 0.98 & 76.52 $\pm$ 0.41 & 85.89 $\pm$ 0.61 & 77.14 $\pm$ 1.01 & 85.10 $\pm$ 0.32 & 75.53 $\pm$ 0.38 \\
MOON     & 83.86 $\pm$ 0.83 & 76.36 $\pm$ 0.64 & 84.93 $\pm$ 0.50 & 76.61 $\pm$ 0.42 & 83.66 $\pm$ 0.19 & 75.17 $\pm$ 0.66 & 82.95 $\pm$ 0.33 & 72.58 $\pm$ 1.50 \\
FedDyn   & 86.08 $\pm$ 0.13 & 72.11 $\pm$ 0.52 & 84.45 $\pm$ 0.36 & 69.88 $\pm$ 2.19 & 85.70 $\pm$ 0.56 & 69.98 $\pm$ 1.10 & 81.01 $\pm$ 0.05 & 65.12 $\pm$ 1.65 \\
SCAFFOLD & 67.77 $\pm$ 3.07 & 60.27 $\pm$ 1.61 & 74.80 $\pm$ 0.77 & 63.46 $\pm$ 1.22 & 73.21 $\pm$ 3.63 & 65.92 $\pm$ 1.42 & 73.45 $\pm$ 3.22 & 66.40 $\pm$ 1.71 \\
FedAdp   & 85.01 $\pm$ 0.79 & 76.73 $\pm$ 1.59 & 85.85 $\pm$ 0.33 & 75.53 $\pm$ 1.01 & 83.89 $\pm$ 3.65 & 76.09 $\pm$ 1.23 & 84.85 $\pm$ 0.66 & 76.28 $\pm$ 0.06 \\
\midrule
& \multicolumn{2}{c}{CHX-5} & \multicolumn{2}{c}{CHX-6}
& \multicolumn{2}{c}{CHX-7} & \multicolumn{2}{c}{} \\
\cmidrule(lr){2-3}\cmidrule(lr){4-5}\cmidrule(lr){6-7}
Algorithm & AUC & GM & AUC & GM & AUC & GM & & \\
\midrule
FedDRAW  & \textbf{80.59 $\pm$ 1.95} & \textbf{73.50 $\pm$ 0.89} & \textbf{83.91 $\pm$ 0.57} & \textbf{73.56 $\pm$ 2.18} & 89.64 $\pm$ 0.37 & \textbf{79.76 $\pm$ 0.13} & & \\
FedAvg   & 79.04 $\pm$ 1.29 & 71.11 $\pm$ 2.59 & 81.53 $\pm$ 1.91 & 71.86 $\pm$ 1.41 & \textbf{89.80 $\pm$ 0.23} & 79.59 $\pm$ 1.07 & & \\
FedNova  & 78.95 $\pm$ 1.03 & 70.18 $\pm$ 0.27 & 81.78 $\pm$ 1.82 & 71.23 $\pm$ 1.63 & 89.16 $\pm$ 1.01 & 79.02 $\pm$ 0.63 & & \\
FedProx  & 79.17 $\pm$ 1.10 & 70.22 $\pm$ 1.93 & 81.71 $\pm$ 0.46 & 69.98 $\pm$ 0.91 & 89.33 $\pm$ 1.08 & 78.29 $\pm$ 0.98 & & \\
MOON     & 79.50 $\pm$ 0.28 & 71.59 $\pm$ 1.72 & 79.21 $\pm$ 1.65 & 66.17 $\pm$ 3.05 & 85.70 $\pm$ 1.05 & 77.15 $\pm$ 0.69 & & \\
FedDyn   & 77.91 $\pm$ 1.05 & 66.07 $\pm$ 6.53 & 78.37 $\pm$ 0.64 & 67.95 $\pm$ 3.63 & 85.48 $\pm$ 0.44 & 75.34 $\pm$ 1.51 & & \\
SCAFFOLD & 69.12 $\pm$ 1.55 & 63.17 $\pm$ 3.70 & 72.74 $\pm$ 1.17 & 65.52 $\pm$ 2.13 & 61.03 $\pm$ 3.56 & 58.85 $\pm$ 3.88 & & \\
FedAdp   & 78.55 $\pm$ 1.47 & 70.47 $\pm$ 1.68 & 81.16 $\pm$ 2.13 & 71.89 $\pm$ 2.62 & 89.71 $\pm$ 0.23 & 79.49 $\pm$ 1.16 & & \\
\bottomrule
\end{tabular}%
}
\end{table}
The largest overall gains arose in CHX-1, where a single large client
faces three small ones. FedDRAW obtained $86.67 \pm 0.25\%$ AUC and
$78.76 \pm 0.18\%$ GM, against $84.98 \pm 0.68\%$ and
$76.01 \pm 1.31\%$ for FedAvg, corresponding to improvements of 1.69
and 2.75 percentage points. The size of this improvement is consistent
with the structure of the partition, since three of the four clients
are small, so a majority of the potentially informative updates are
down-weighted under proportional aggregation, and correcting this
under-representation affects a correspondingly large share of the
federation. In addition, the standard deviations of FedDRAW were
markedly smaller than those of FedAvg, indicating that redistributing
influence away from a single dominant client also reduced run-to-run
variability.

The intermediate settings CHX-2 and CHX-3 followed the same pattern. In
CHX-2, FedDRAW reached $87.46 \pm 0.30\%$ AUC and $77.83 \pm 0.81\%$
GM, improving over FedAvg ($85.77 \pm 0.33\%$ and $75.41 \pm 1.32\%$)
by 1.69 and 2.42 percentage points. In CHX-3, FedDRAW achieved
$86.64 \pm 0.18\%$ AUC and $78.03 \pm 0.91\%$ GM, compared with
$85.71 \pm 0.43\%$ and $76.71 \pm 1.80\%$ for FedAvg, corresponding to
gains of 0.93 and 1.32 percentage points. Taken together, CHX-1 to
CHX-4 show that the advantage of FedDRAW grows with the fraction of the
federation that is under-represented by proportional weighting, and
that the method remains effective even in the most asymmetric
configuration.

The class-enriched scenarios CHX-5 and CHX-6 provide a more targeted
assessment under distribution heterogeneity. FedDRAW assigns a single
aggregation weight to each client and performs no class-wise
aggregation; the purpose of these scenarios is therefore not to
determine whether FedDRAW identifies an individually dominant
pathology, but whether the overall model update of a small yet
informative client can contribute effectively to global optimization
despite its limited local dataset. In CHX-5, FedDRAW achieved
$80.59 \pm 1.95\%$ AUC and $73.50 \pm 0.89\%$ GM, improving over FedAvg
($79.04 \pm 1.29\%$ and $71.11 \pm 2.59\%$) by 1.55 and 2.39 percentage
points; in CHX-6, FedDRAW reached $83.91 \pm 0.57\%$ AUC and
$73.56 \pm 2.18\%$ GM, compared with $81.53 \pm 1.91\%$ and
$71.86 \pm 1.41\%$ for FedAvg, corresponding to gains of 2.38 and 1.70
percentage points. In both scenarios, no competing method exceeded
FedDRAW in either metric. These improvements can be attributed to the
fact that the enriched client contributes gradient information for a
pathology that is comparatively scarce at the large clients; under
proportional weighting this information is diluted by a factor
determined solely by sample count, whereas FedDRAW allows it to enter
the global model with greater influence during the rounds in which it
is most useful. Notably, the enriched pathology constitutes only one
component of the small client's multi-label data, and its local update
is shaped by the complete joint label distribution, which also includes
the remaining four target pathologies. The results therefore support
the client-level design of FedDRAW rather than a class-specific
interpretation of its weighting mechanism.

Finally, the homogeneous reference scenario CHX-7 confirms that the
adaptive mechanism is safe when no meaningful heterogeneity is present.
FedDRAW obtained the highest GM of $79.76 \pm 0.13\%$, while its AUC of
$89.64 \pm 0.37\%$ lay within 0.16 percentage points of the best value
in this scenario. Here again the two metrics do not agree on the
ordering of the methods, although the clients are of equal size and no
method has a structural advantage. The proportional weights of FedAvg
are already close to uniform in this setting, so there is little
imbalance for the adaptive mechanism to compensate; the relevant
observation is that FedDRAW retains the leading value under the metric
that reflects the deployed operating points, and that the adaptive
weighting incurs no performance penalty when its principal advantage is
not required.

\begin{table}[!htbp]
\centering
\refstepcounter{table}
\label{tab:results_chm}
\noindent\parbox{\textwidth}{%
\scriptsize
\textbf{Table \thetable.} Performance (\%) on ChestMNIST across the five
client-partition scenarios (mean $\pm$ standard deviation over five runs).
Best values per column are shown in bold.
}
\vspace{4pt}
\resizebox{\textwidth}{!}{%
\begin{tabular}{lcccccccccc}
\toprule
& \multicolumn{2}{c}{CHM-1} & \multicolumn{2}{c}{CHM-2}
& \multicolumn{2}{c}{CHM-3} & \multicolumn{2}{c}{CHM-4}
& \multicolumn{2}{c}{CHM-5} \\
\cmidrule(lr){2-3}\cmidrule(lr){4-5}\cmidrule(lr){6-7}\cmidrule(lr){8-9}\cmidrule(lr){10-11}
Algorithm & AUC & GM & AUC & GM & AUC & GM & AUC & GM & AUC & GM \\
\midrule
FedDRAW  & \textbf{81.78 $\pm$ 0.53} & \textbf{74.84 $\pm$ 0.29} & \textbf{82.91 $\pm$ 0.13} & \textbf{75.86 $\pm$ 0.40} & 84.18 $\pm$ 0.14 & \textbf{76.61 $\pm$ 0.34} & \textbf{82.67 $\pm$ 0.77} & \textbf{75.48 $\pm$ 0.42} & 83.86 $\pm$ 0.69 & 76.53 $\pm$ 0.37 \\
FedAvg   & 80.13 $\pm$ 0.17 & 73.06 $\pm$ 0.16 & 80.14 $\pm$ 0.60 & 73.22 $\pm$ 0.45 & 84.10 $\pm$ 0.27 & 76.01 $\pm$ 0.28 & 81.69 $\pm$ 0.21 & 73.67 $\pm$ 0.41 & 83.97 $\pm$ 0.10 & 76.30 $\pm$ 0.14 \\
FedNova  & 78.36 $\pm$ 1.87 & 71.77 $\pm$ 1.49 & 80.69 $\pm$ 0.12 & 73.26 $\pm$ 0.11 & \textbf{84.25 $\pm$ 0.21} & 75.62 $\pm$ 0.06 & 81.37 $\pm$ 0.34 & 74.18 $\pm$ 0.09 & 83.77 $\pm$ 0.13 & 76.24 $\pm$ 0.15 \\
FedProx  & 80.33 $\pm$ 0.67 & 73.16 $\pm$ 0.28 & 80.96 $\pm$ 0.55 & 73.58 $\pm$ 0.52 & 84.13 $\pm$ 0.25 & 75.72 $\pm$ 0.25 & 81.22 $\pm$ 0.68 & 72.12 $\pm$ 1.04 & \textbf{84.22 $\pm$ 0.04} & 76.61 $\pm$ 0.06 \\
MOON     & 80.37 $\pm$ 0.41 & 73.04 $\pm$ 0.50 & 78.60 $\pm$ 0.51 & 71.40 $\pm$ 0.34 & 80.03 $\pm$ 0.60 & 72.95 $\pm$ 0.59 & 79.51 $\pm$ 0.43 & 72.50 $\pm$ 0.37 & 82.27 $\pm$ 0.16 & 74.99 $\pm$ 0.23 \\
FedDyn   & 72.99 $\pm$ 0.80 & 67.53 $\pm$ 0.41 & 71.79 $\pm$ 0.35 & 66.68 $\pm$ 0.11 & 74.63 $\pm$ 1.37 & 68.57 $\pm$ 1.09 & 75.74 $\pm$ 0.54 & 72.39 $\pm$ 1.88 & 74.86 $\pm$ 0.85 & 68.89 $\pm$ 0.65 \\
SCAFFOLD & 63.36 $\pm$ 0.92 & 59.83 $\pm$ 0.73 & 61.62 $\pm$ 1.39 & 58.31 $\pm$ 1.52 & 68.14 $\pm$ 1.03 & 63.36 $\pm$ 0.81 & 65.23 $\pm$ 1.37 & 62.70 $\pm$ 0.63 & 60.56 $\pm$ 3.87 & 57.45 $\pm$ 2.99 \\
FedAdp   & 80.25 $\pm$ 0.48 & 72.95 $\pm$ 0.36 & 80.20 $\pm$ 0.43 & 73.85 $\pm$ 0.45 & 83.97 $\pm$ 0.24 & 76.15 $\pm$ 0.37 & 81.26 $\pm$ 0.41 & 73.64 $\pm$ 2.01 & 83.83 $\pm$ 0.10 & \textbf{76.66 $\pm$ 0.14} \\
\bottomrule
\end{tabular}%
}
\end{table}

\FloatBarrier
% ---------------------------------------------------------------------
\subsection{Results on ChestMNIST}
\label{sec:results_chestmnist}

Table~\ref{tab:results_chm} reports the ChestMNIST results under the
five scenarios. Because ChestMNIST is the noisier of the two datasets,
it permits client partitions with more pronounced heterogeneity and
therefore allows FedDRAW to be examined under more severe conditions
than CheXpert. FedDRAW attained the highest GM in the four
heterogeneous configurations, while the AUC differences between the
leading methods were small; this pattern is examined in detail below
and is itself informative about the relative sensitivity of the two
metrics.

In CHM-1, which represents a pronounced two-client quantity imbalance,
FedDRAW achieved $81.78 \pm 0.53\%$ AUC and $74.84 \pm 0.29\%$ GM,
compared with $80.13 \pm 0.17\%$ and $73.06 \pm 0.16\%$ for FedAvg,
corresponding to improvements of 1.65 and 1.78 percentage points. The
advantage widened further in CHM-2, in which FedDRAW obtained
$82.91 \pm 0.13\%$ AUC and $75.86 \pm 0.40\%$ GM, against
$80.14 \pm 0.60\%$ and $73.22 \pm 0.45\%$ for FedAvg, corresponding to
improvements of 2.77 and 2.64 percentage points; in both scenarios, no
competing method reached a higher value of either metric. As in the
corresponding CheXpert experiments, the gains were largest when small
clients collectively held a substantial fraction of the informative
updates, supporting the interpretation that the benefit of FedDRAW
stems from counteracting the systematic dilution of small-client
contributions rather than from a dataset-specific effect.

FedDRAW also achieved the best performance in the class-enriched
scenario CHM-4, in which the smaller client was intentionally enriched
for Atelectasis while retaining examples of the other target
pathologies. FedDRAW reached an AUC of $82.67 \pm 0.77\%$ and a GM of
$75.48 \pm 0.42\%$, compared with $81.69 \pm 0.21\%$ and
$73.67 \pm 0.41\%$ for FedAvg, corresponding to improvements of 0.98
and 1.81 percentage points. Together with the corresponding CheXpert
experiments, this finding indicates that the benefit of adaptive client
weighting is not restricted to a single dataset and is consistent with
the hypothesis that a client's contribution to federated optimization
cannot always be adequately represented by its local sample count
alone.

CHM-3 is the most severe configuration, in which the small client holds
only $10.8\%$ of the data in a six-client federation and the five large
clients jointly dominate the sample count. In this scenario, FedDRAW
attained the highest GM of $76.61 \pm 0.34\%$ among all methods, while
its AUC of $84.18 \pm 0.14\%$ lay $0.07$ percentage points below the
strongest AUC ($84.25 \pm 0.21\%$), a difference well within one
standard deviation. This is precisely the situation anticipated in
Section~\ref{sec:evaluation_metrics}. Under strong class imbalance
combined with client heterogeneity, AUC is not sufficient to
characterize a classifier, because it is invariant to prevalence and
aggregates over thresholds that are never used in practice. Methods
that appear indistinguishable, or even marginally preferable, under AUC
are separated once the operating points are taken into account, and it
is at those operating points that FedDRAW realizes the benefit of
reallocating influence toward the informative small client. The result
therefore confirms both the necessity of reporting GM in this setting
and the effectiveness of FedDRAW under the most severe heterogeneity
considered here.

The homogeneous reference scenario CHM-5 confirms the same conclusion
from the opposite direction. Here the ordering of the methods under AUC
again differs from the ordering under GM, even though all clients hold
equal amounts of data and no method has a structural advantage. Taken
together with CHM-3, this shows that AUC alone cannot decide between
aggregation strategies on imbalanced multi-label chest radiographs,
since the metric is invariant to prevalence and averages over
thresholds that are never used, so it neither exposes the disadvantage
of proportional weighting under heterogeneity nor credits the
correction that FedDRAW applies. GM, which evaluates the classifiers at
the thresholds under which they would be deployed, is therefore the
metric that reflects the objective of this work.
% ---------------------------------------------------------------------
\subsection{Statistical analysis}
\label{sec:statistical_analysis}

To assess whether the observed differences between the algorithms are
systematic across the full experimental design, we used the Friedman
test~\citep{demsar2006statistical}, a non-parametric test that compares
multiple algorithms over multiple datasets without assuming normality
or homogeneity of variances. Because AUC and GM capture two different
properties of the trained models, we analyzed the two metrics
separately. For each metric, the eight methods were ranked within every
scenario, with rank 1 denoting the best method and tied methods
receiving the average of the tied ranks, yielding $N = 12$ ranking
blocks over $k = 8$ algorithms per metric. Table~\ref{tab:ranks}
reports the AUC and GM ranks of every method in every scenario together
with the per-metric average ranks.

The Friedman test rejected the null hypothesis of equal performance for
both metrics ($\chi^2_F(7) = 58.61$ for AUC and $\chi^2_F(7) = 63.50$
for GM, both $p < 0.001$), confirming that the choice of aggregation
strategy has a significant effect on performance. FedDRAW obtained the
best average rank for both metrics, 1.50 for AUC and 1.17 for GM,
ranking first in eight of the twelve scenarios for AUC and in eleven of
the twelve for GM; the next-best average ranks were 2.92 for AUC and
3.08 for GM.

After establishing significant differences with the Friedman test, we
performed the Nemenyi post-hoc test~\citep{demsar2006statistical}. If
the difference in average ranks between two algorithms exceeds a
critical difference (CD), the difference is significant. Given the
moderate number of scenarios ($N = 12$) relative to the number of
compared algorithms ($k = 8$), we used a significance level of
$\alpha = 0.10$, for which the critical value $q_\alpha$ for $k = 8$
algorithms is $2.780$~\citep{demsar2006statistical}. Eq.~\ref{eq:cd} is
the result of our calculation of the critical difference.
\begin{equation}
\mathrm{CD} \;=\; q_\alpha \sqrt{\frac{k(k+1)}{6N}} \;=\; 2.78 ,
\label{eq:cd}
\end{equation}

This indicates that any pairwise difference in average ranks larger
than 2.78 is statistically significant. On AUC, FedDRAW performed
significantly better than FedAdp~\citep{wu2021fedadp},
MOON~\citep{li2021moon}, FedDyn~\citep{acar2021feddyn}, and
SCAFFOLD~\citep{karimireddy2020scaffold}. On GM, FedDRAW performed
significantly better than FedProx~\citep{li2020fedprox},
FedNova~\citep{wang2020fednova}, MOON~\citep{li2021moon},
FedDyn~\citep{acar2021feddyn}, and
SCAFFOLD~\citep{karimireddy2020scaffold}. The stronger separation
obtained under GM is consistent with the design of FedDRAW, whose
reallocation of aggregation weight primarily improves the
sensitivity--specificity balance at the operating thresholds.
Figure~\ref{fig:nemenyi} visualizes both analyses as
critical-difference diagrams.

\begin{table}[!htbp]
\centering

\refstepcounter{table}
\label{tab:ranks}

\noindent\parbox{\textwidth}{%
\scriptsize
\textbf{Table \thetable.} Per-scenario ranks of every method used in the
Friedman test (AUC\,/\,GM; rank 1 = best, tied methods receive the average
of the tied ranks). The last two rows report the average AUC and GM ranks
across the twelve scenarios.
}

\vspace{3pt}

\scriptsize
\setlength{\tabcolsep}{3pt}
\renewcommand{\arraystretch}{0.88}

\begin{tabular}{lcccccccc}
\toprule
Scenario & FedDRAW & FedAvg & FedNova & FedProx & MOON & FedDyn & SCAFFOLD & FedAdp \\
\midrule
CHX-1 & 1\,/\,1 & 6\,/\,6 & 4\,/\,3 & 3\,/\,5 & 7\,/\,4 & 2\,/\,7 & 8\,/\,8 & 5\,/\,2 \\
CHX-2 & 1\,/\,1 & 5\,/\,6 & 2\,/\,4 & 3\,/\,3 & 6\,/\,2 & 7\,/\,7 & 8\,/\,8 & 4\,/\,5 \\
CHX-3 & 1\,/\,1 & 3\,/\,4 & 5\,/\,3 & 2\,/\,2 & 7\,/\,6 & 4\,/\,7 & 8\,/\,8 & 6\,/\,5 \\
CHX-4 & 2\,/\,1 & 1\,/\,4 & 6\,/\,3 & 3\,/\,5 & 5\,/\,6 & 7\,/\,8 & 8\,/\,7 & 4\,/\,2 \\
CHX-5 & 1\,/\,1 & 4\,/\,3 & 5\,/\,6 & 3\,/\,5 & 2\,/\,2 & 7\,/\,7 & 8\,/\,8 & 6\,/\,4 \\
CHX-6 & 1\,/\,1 & 4\,/\,3 & 2\,/\,4 & 3\,/\,5 & 6\,/\,7 & 7\,/\,6 & 8\,/\,8 & 5\,/\,2 \\
CHX-7 & 3\,/\,1 & 1\,/\,2 & 5\,/\,4 & 4\,/\,5 & 6\,/\,6 & 7\,/\,7 & 8\,/\,8 & 2\,/\,3 \\
CHM-1 & 1\,/\,1 & 5\,/\,3 & 6\,/\,6 & 3\,/\,2 & 2\,/\,4 & 7\,/\,7 & 8\,/\,8 & 4\,/\,5 \\
CHM-2 & 1\,/\,1 & 5\,/\,5 & 3\,/\,4 & 2\,/\,3 & 6\,/\,6 & 7\,/\,7 & 8\,/\,8 & 4\,/\,2 \\
CHM-3 & 2\,/\,1 & 4\,/\,3 & 1\,/\,5 & 3\,/\,4 & 6\,/\,6 & 7\,/\,7 & 8\,/\,8 & 5\,/\,2 \\
CHM-4 & 1\,/\,1 & 2\,/\,3 & 3\,/\,2 & 5\,/\,7 & 6\,/\,5 & 7\,/\,6 & 8\,/\,8 & 4\,/\,4 \\
CHM-5 & 3\,/\,3 & 2\,/\,4 & 5\,/\,5 & 1\,/\,2 & 6\,/\,6 & 7\,/\,7 & 8\,/\,8 & 4\,/\,1 \\
\midrule
Average rank (AUC) & \textbf{1.50} & 3.50 & 3.92 & 2.92 & 5.42 & 6.33 & 8.00 & 4.42 \\
Average rank (GM)  & \textbf{1.17} & 3.83 & 4.08 & 4.00 & 5.00 & 6.92 & 7.92 & 3.08 \\
\bottomrule
\end{tabular}

\end{table}

\vspace{3pt}

\begin{center}

\begin{minipage}[c]{0.48\textwidth}
    \centering
    \includegraphics[width=0.92\linewidth]{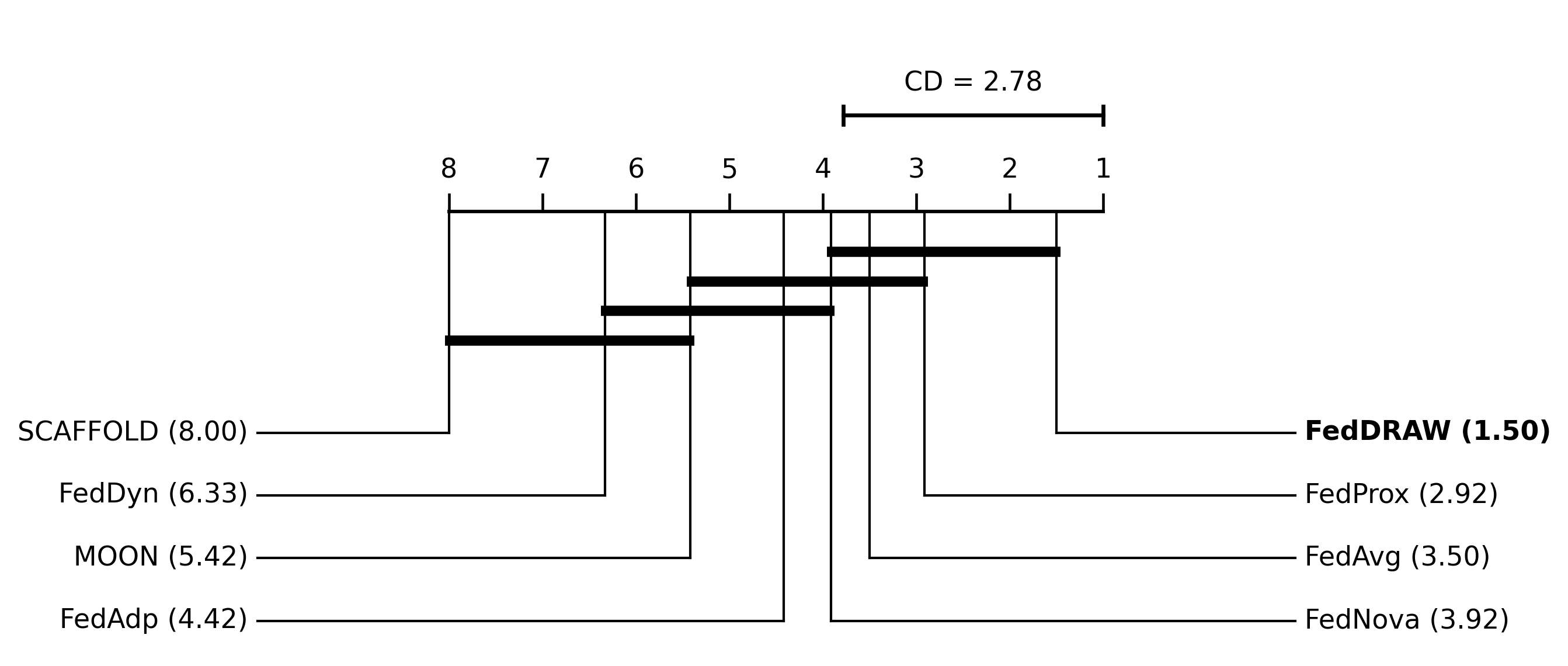}
\end{minipage}
\hfill
\begin{minipage}[c]{0.48\textwidth}
    \centering
    \includegraphics[width=0.92\linewidth]{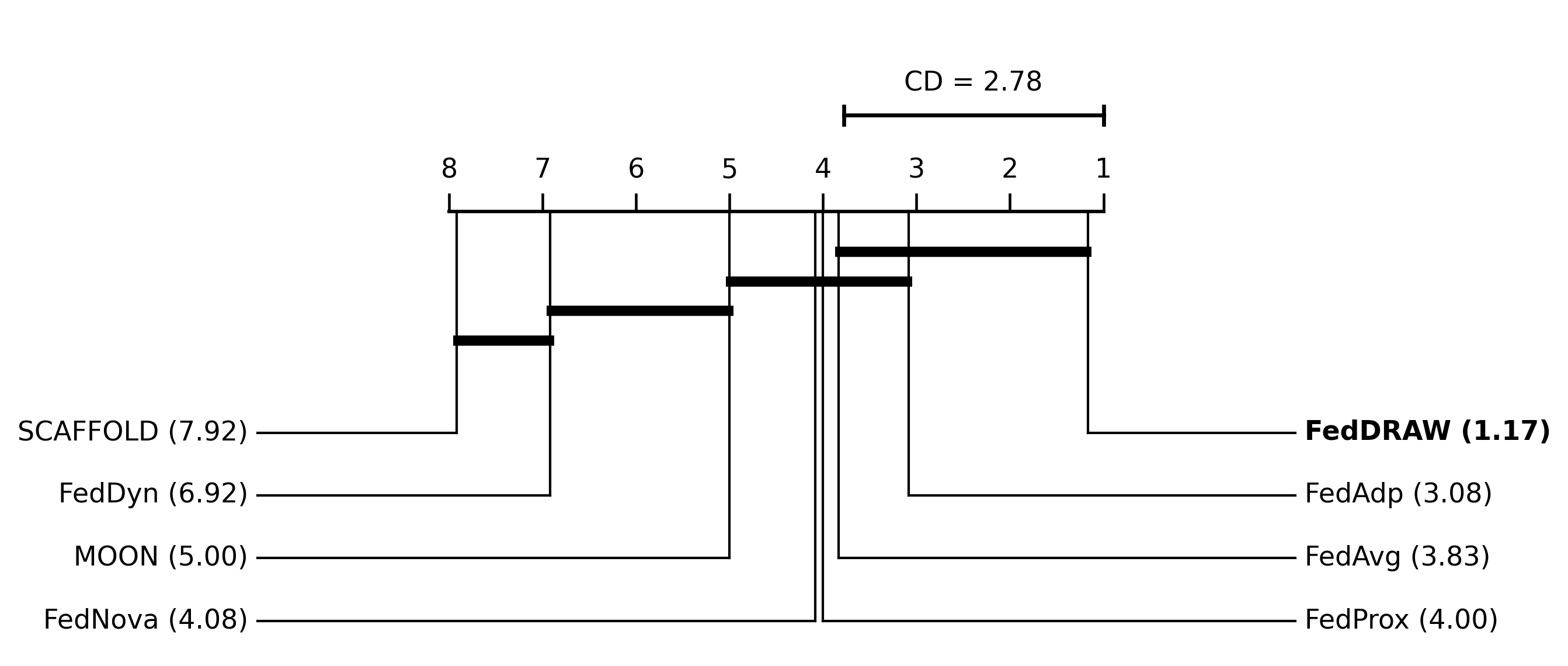}
\end{minipage}

\vspace{2pt}

\refstepcounter{figure}
\label{fig:nemenyi}

\noindent\parbox{\textwidth}{%
\scriptsize
\textbf{Figure \thefigure.} Critical-difference diagrams of the Nemenyi
post-hoc test over the twelve scenarios for AUC (left) and GM (right)
($\alpha = 0.10$, $\mathrm{CD} = 2.78$). Algorithms connected by a
horizontal bar are not significantly different.
}

\end{center}

\vspace{4pt}

% ---------------------------------------------------------------------
\subsection{Analysis of FedDRAW aggregation dynamics}
\label{sec:aggregation_dynamics}

The central design objective of FedDRAW is that a small client whose
updates are informative should not be permanently constrained to the
contribution implied by its sample count. Figure~\ref{fig:weights}
examines this behavior in the two CheXpert scenarios with the most
pronounced size asymmetry.

In CHX-4, Clients 1--9 each hold 10{,}000 samples and Client 10 holds
2{,}500, so that Client 10 represents approximately $2.7\%$ of the
training data against approximately $10.8\%$ for each larger client.
Under sample-size-proportional aggregation this ratio would be fixed
for the entire run, and FedDRAW indeed begins from exactly this
allocation, assigning Client 10 an initial weight of $0.0270$ because
the round-0 weights are given by the data-size prior alone. Within four
rounds, the weight of Client 10 rises above the uniform level
$1/K = 0.10$ and reaches a maximum of $0.1037$ in round~6, at which
point the small client receives a larger share of the aggregate than
any of the four times larger clients
(Figure~\ref{fig:weights}a). The weight then declines and reaches
$0.1000$ by round~14. The reallocation is therefore not a transient
artifact of initialization but a sustained reweighting over seven
consecutive communication rounds, driven by the inner schedule as
reliance shifts from the data-size prior toward last-layer similarity,
and subsequently released by the outer schedule.

CHX-1 exhibits the same mechanism from the opposite direction. A single
client of 20{,}000 samples faces three clients of 5{,}000 samples each
and would hold a fixed weight of $0.5714$ under proportional
aggregation. FedDRAW starts from this same weight and reduces it
rapidly; the three small clients overtake it in round~5, and the large
client reaches its minimum of $0.2334$ in round~6 before returning to
$0.2500$ as the outer schedule drives the weights toward uniformity
(Figure~\ref{fig:weights}c). The initial dominance is thus dismantled
within five rounds, consistent with the pronounced improvement observed
for this configuration (Section~\ref{sec:results_chexpert}).

\begin{center}

\begin{minipage}[t]{0.49\textwidth}
  \centering
  \includegraphics[width=\linewidth]{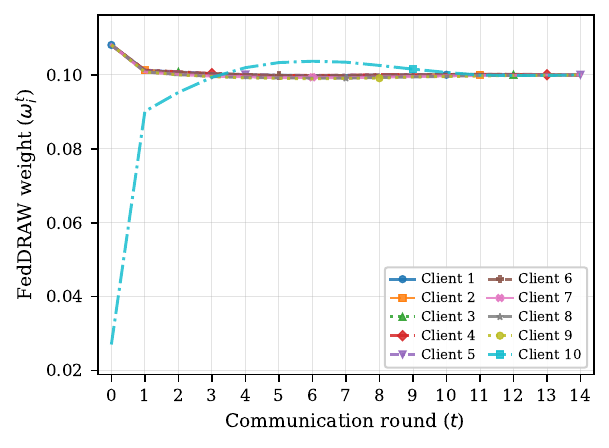}

  \vspace{-2pt}
  {\scriptsize\textbf{(a)}}
\end{minipage}
\hfill
\begin{minipage}[t]{0.49\textwidth}
  \centering
  \includegraphics[width=\linewidth]{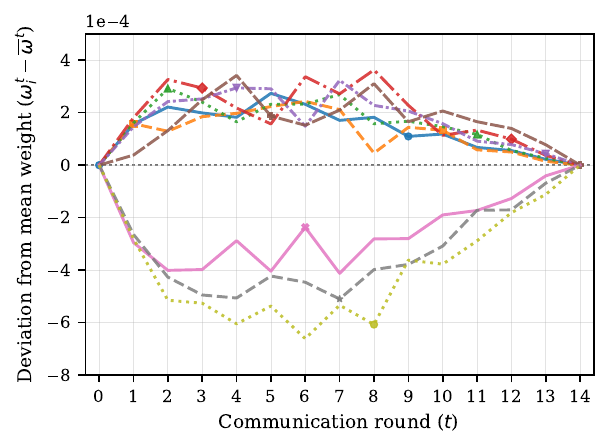}

  \vspace{-2pt}
  {\scriptsize\textbf{(b)}}
\end{minipage}

\vspace{0.4em}

\begin{minipage}[t]{0.49\textwidth}
  \centering
  \includegraphics[width=\linewidth]{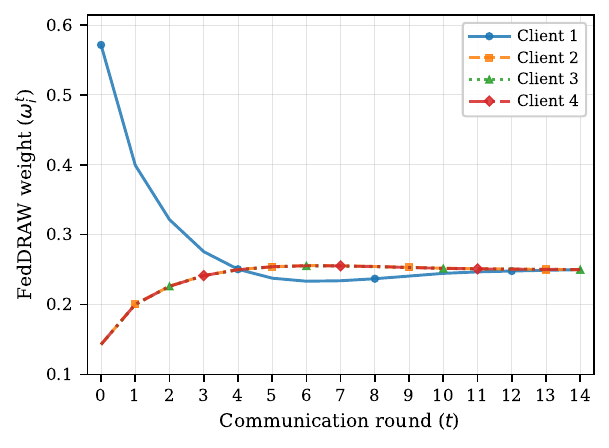}

  \vspace{-2pt}
  {\scriptsize\textbf{(c)}}
\end{minipage}
\hfill
\begin{minipage}[t]{0.49\textwidth}
  \centering
  \includegraphics[width=\linewidth]{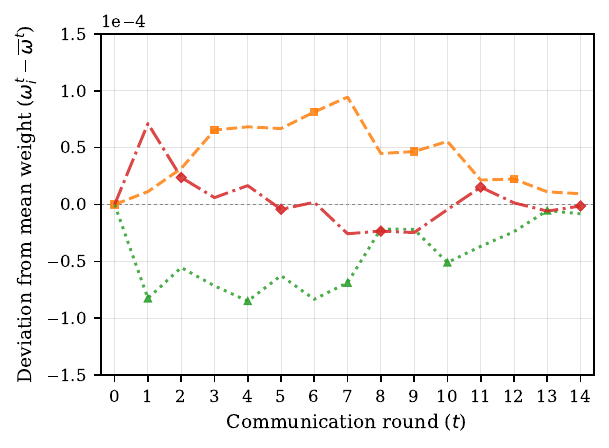}

  \vspace{-2pt}
  {\scriptsize\textbf{(d)}}
\end{minipage}

\vspace{4pt}

\refstepcounter{figure}
\label{fig:weights}

\noindent\parbox{\textwidth}{%
\scriptsize
\textbf{Figure \thefigure.}
Evolution of FedDRAW aggregation weights $\omega_i^t$ across
communication rounds on CheXpert. (a)~CHX-4, Clients 1--9 with 10{,}000
samples each and Client 10 with 2{,}500 samples; the weight of the small
client rises above the uniform level and exceeds that of every larger
client between rounds~4 and~10. (b)~Deviation of Clients 1--9 from
their mean weight in CHX-4. (c)~CHX-1, Client 1 with 20{,}000 samples
and Clients 2--4 with 5{,}000 samples each; the initial dominance of the
large client is dismantled within five rounds. (d)~Deviation of Clients
2--4 from their mean weight in CHX-1. Panels~(b) and~(d) show that
equally sized clients are differentiated during training on a scale that
is not resolvable in panels~(a) and~(c). All panels share the client
colors and line styles given in the legends of panels~(a) and~(c).
}

\end{center}

On the scale of Figures~\ref{fig:weights}a and~\ref{fig:weights}c the
curves of the equally sized clients appear to overlap, but they are not
identical; panels~(b) and~(d) therefore plot their deviation from the
group mean, which resolves differences that the common scale cannot
display. The curves are distinct rather than degenerate, since FedDRAW
separates same-size clients on the order of $10^{-4}$, and in CHX-4 the
separation is systematic, with Clients 7--9 remaining below the mean in
every round. The spread reaches $9.98 \times 10^{-4}$ in round~6 in
CHX-4 and $3.60 \times 10^{-4}$ in round~4 in CHX-1, and vanishes
exactly in the final round of both scenarios, where $\beta_{T-1} = 0$
enforces uniform weights. Equally sized clients are therefore
distinguished while the similarity signal is informative and equalized
once the outer schedule has annealed. These dynamics should be read together with the predictive results. In
CHX-4, FedDRAW achieved $85.57 \pm 0.60\%$ AUC and $77.35 \pm 0.13\%$
GM against $85.76 \pm 0.68\%$ and $75.61 \pm 0.27\%$ for FedAvg, and
the reallocation of weight toward the small client improved the
sensitivity--specificity balance by 1.74 percentage points of GM at
essentially unchanged discrimination. Overall, the experiments demonstrate that the advantage of FedDRAW is
most evident under heterogeneous client configurations, particularly
when comparatively small clients contain useful information that would
receive limited influence under conventional sample-size-based
aggregation.

\FloatBarrier

%=====================================================================
% CONCLUSION
%=====================================================================
\section{Conclusion}
\label{sec:conclusion}

In this work, we addressed a structural limitation of
sample-size-proportional aggregation in federated learning. Weighting
clients by their number of local samples means that influence is
granted to the largest participants by construction, although a smaller
client may hold informative samples, including presentations of a
pathology that are scarce or entirely absent at the larger clients.
 Moreover, because all clients continue to train throughout the federation, a weight that is fixed in advance cannot reflect how informative a client's updates actually are at a given stage of training. We proposed FedDRAW, a server-side aggregation strategy that
blends the cosine similarity between client and global classification
parameters with a data-size prior under two coupled exponential
annealing schedules. The resulting weights follow a deliberate
trajectory over training, in which the size prior anchors aggregation in the early rounds, model similarity gains influence as training advances, and a decaying temperature drives all weights toward uniformity as training converges. FedDRAW leaves client-side training
unchanged, introduces no additional communication, and was applied with
a single fixed hyperparameter configuration across all experiments.

Across twelve client-partition scenarios on CheXpert and ChestMNIST,
FedDRAW achieved the highest GM in all seven CheXpert configurations
and the highest AUC in five of them, as well as the highest values of
both metrics in the principal heterogeneous ChestMNIST configurations,
with the largest gains arising precisely where proportional weighting
is most restrictive, that is, when a substantial fraction of the
federation consists of small or class-enriched clients. In homogeneous
settings, where little imbalance exists to compensate, FedDRAW remained
competitive with the strongest baselines, indicating that the adaptive
mechanism incurs no penalty when its principal advantage is not
required. The analysis of the aggregation dynamics confirmed that the
mechanism operates as designed, since a client holding only 2.7\% of
the training data transiently received an above-proportional share of
influence before the weights converged toward uniformity. Our
evaluation further showed that the relative ranking of federated
methods can differ between threshold-independent discrimination and
threshold-dependent operating performance, supporting the practice of
reporting GM alongside AUC for imbalanced, noisily labeled chest
radiograph data.

As future work, we plan to extend FedDRAW from client-level to
class-level adaptivity. In its present form, the method assigns a single
scalar weight to each client, so a client that is informative for one
pathology but not for others still receives one aggregate weight. A
class-wise variant, in which the aggregation weight is computed
separately for each pathology, would allow every client to contribute
selectively to the findings for which its data are most informative, and
is a natural direction for extending the reputation-based weighting
proposed here.

\section*{Acknowledgment}

This work was supported by BMFTR (grants 01D2208A and
01KD2414A/FAIrPaCT), the hessian.AI Innovation Lab
(Hessian Ministry for Digital Strategy and Innovation, grant
S-DIW04/0013/003), and NHR-Nord@Göttingen (Emmy/Grete) as part of the
NHR infrastructure, partly funded by the Deutsche Forschungsgemeinschaft
(DFG, grant 405797229).

\bibliographystyle{cas-model2-names}
\bibliography{feddraw_references}

\end{document}